\documentclass{article}
\pdfoutput=1
\usepackage{times}
\usepackage{graphicx} 
\usepackage{subfigure} 
\usepackage{natbib}
\usepackage{algorithm}
\usepackage{algorithmic}
\usepackage{hyperref}

\usepackage[accepted]{icml2015_copy}
\usepackage{array}
\usepackage{booktabs}
\usepackage{tikz}
\newcommand{\ii}{{\bf i}}
\newcommand{\cc}{{\bf c}}
\newcommand{\oo}{{\bf o}}
\newcommand{\ff}{{\bf f}}
\newcommand{\hh}{{\bf h}}
\newcommand{\xx}{{\bf x}}

\newcommand{\bb}{{\bf b}}

\newcommand{\Figref}[1]{Fig.~\ref{#1}}
\newcommand{\Tabref}[1]{Table~\ref{#1}}
\newcommand{\Secref}[1]{Sec.~\ref{#1}}
\newcolumntype{L}[1]{>{\raggedright\let\newline\\\arraybackslash\hspace{0pt}}m{#1}}
\newcolumntype{C}[1]{>{\centering\let\newline\\\arraybackslash\hspace{0pt}}m{#1}}
\newcolumntype{R}[1]{>{\raggedleft\let\newline\\\arraybackslash\hspace{0pt}}m{#1}}

\icmltitlerunning{Unsupervised Learning with LSTMs}

\begin{document} 
\twocolumn[
\icmltitle{Unsupervised Learning of Video Representations using LSTMs}

\icmlauthor{Nitish Srivastava}{nitish@cs.toronto.edu}
\icmlauthor{Elman Mansimov}{emansim@cs.toronto.edu}
\icmlauthor{Ruslan Salakhutdinov}{rsalakhu@cs.toronto.edu}
\icmladdress{University of Toronto,
             6 Kings College Road, Toronto, ON M5S 3G4 CANADA}
\icmlkeywords{unsupervised learning, deep learning, sequence learning, video, action recognition.}
\vskip 0.3in
]

\begin{abstract}

We use multilayer Long Short Term Memory (LSTM) networks to learn
representations of video sequences. Our model uses an encoder LSTM to map an
input sequence into a fixed length representation. This representation is
decoded using single or multiple decoder LSTMs to perform different tasks, such
as reconstructing the input sequence, or predicting the future sequence. We
experiment with two kinds of input sequences -- patches of image pixels and
high-level representations (``percepts") of video frames extracted using a
pretrained convolutional net.  We explore different design choices such as
whether the decoder LSTMs should condition on the generated output. We analyze the
outputs of the model qualitatively to see how well the model can extrapolate the
learned video representation into the future and into the past. We try to
visualize and interpret the learned features. We stress test the model by
running it on longer time scales and on out-of-domain data. We further evaluate
the representations by finetuning them for a supervised learning problem --
human action recognition on the UCF-101 and HMDB-51 datasets. We show that the
representations help improve classification accuracy, especially when there are
only a few training examples.  Even models pretrained on unrelated datasets (300
hours of YouTube videos) can help action recognition performance.
\end{abstract} 

\section{Introduction}
\label{submission}

Understanding temporal sequences is important for solving many problems in the
AI-set. Recently, recurrent neural networks using the Long Short Term Memory
(LSTM) architecture \citep{Hochreiter} have been used successfully to perform various supervised
sequence learning tasks, such as speech recognition \citep{graves_lstm}, machine
translation \citep{IlyaMT, ChoMT}, and caption generation for images
\citep{OriolCaption}. They have also been applied on videos for recognizing
actions and generating natural language descriptions \citep{BerkeleyVideo}. A
general sequence to sequence learning framework was described by \citet{IlyaMT}
in which a recurrent network is used to encode a sequence into a fixed length
representation, and then another recurrent network is used to decode a sequence
out of that representation. In this work, we apply and extend this framework to
learn representations of sequences of images. We choose to work in the
\emph{unsupervised} setting where we only have access to a dataset of unlabelled
videos.

Videos are an abundant and rich source of visual information and can be seen as
a window into the physics of the world we live in, showing us examples of what
constitutes objects, how objects move against backgrounds, what happens when
cameras move and how things get occluded. Being able to learn a representation
that disentangles these factors would help in making intelligent machines that
can understand and act in their environment.  Additionally, learning good video
representations is essential for a number of useful tasks, such as recognizing
actions and gestures.

\subsection{Why Unsupervised Learning?}
Supervised learning has been extremely successful in learning good visual
representations that not only produce good results at the task they are trained
for, but also transfer well to other tasks and datasets. Therefore, it is
natural to extend the same approach to learning video representations. This has
led to research in 3D convolutional nets \citep{conv3d, C3D}, different temporal
fusion strategies \citep{KarpathyCVPR14} and exploring different ways of
presenting visual information to convolutional nets \citep{Simonyan14b}.
However, videos are much higher dimensional entities compared to single images.
Therefore, it becomes increasingly difficult to do credit assignment and learn long
range structure, unless we collect much more labelled data or do a lot of
feature engineering (for example computing the right kinds of flow features) to
keep the dimensionality low. The costly work of collecting more labelled data
and the tedious work of doing more clever engineering can go a long way in
solving particular problems, but this is ultimately unsatisfying as a machine
learning solution. This highlights the need for using unsupervised learning to
find and represent structure in videos.  Moreover, videos have a lot of
structure in them (spatial and temporal regularities) which makes them
particularly well suited as a domain for building unsupervised learning models.

\subsection{Our Approach}
When designing any unsupervised learning model, it is crucial to have the right
inductive biases and choose the right objective function so that the learning
signal points the model towards learning useful features. In
this paper, we use the LSTM Encoder-Decoder framework to learn video
representations. The key inductive bias here is that the same operation must be
applied at each time step to propagate information to the next step. This
enforces the fact that the physics of the world remains the same, irrespective of
input. The same physics acting on any state, at any time, must produce the next
state. Our model works as follows.
The Encoder LSTM runs through a sequence of frames to come up
with a representation. This representation is then decoded through another LSTM
to produce a target sequence.  We consider different choices of the target
sequence.  One choice is to predict the same sequence as the input. The
motivation is similar to that of autoencoders -- we wish to capture all that is
needed to reproduce the input but at the same time go through the inductive
biases imposed by the model. Another option is to predict the future frames.
Here the motivation is to learn a representation that extracts all that is
needed to extrapolate the motion and appearance beyond what has been observed. These two
natural choices can also be combined. In this case, there are two decoder LSTMs
-- one that decodes the representation into the input sequence and another that
decodes the same representation to predict the future.

The inputs to the model can, in principle, be any representation of individual
video frames. However, for the purposes of this work, we limit our attention to
two kinds of inputs. The first is image patches. For this we use natural image
patches as well as a dataset of moving MNIST digits. The second is
high-level ``percepts" extracted by applying a convolutional net trained on
ImageNet. These percepts are the states of last (and/or second-to-last) layers of
rectified linear hidden states from a convolutional neural net model.

In order to evaluate the learned representations we qualitatively analyze the
reconstructions and predictions made by the model. For a more quantitative
evaluation, we use these LSTMs as initializations for the supervised task of
action recognition. If the unsupervised learning model comes up with useful
representations then the classifier should be able to perform better, especially
when there are only a few labelled examples. We find that this is indeed the
case.

\subsection{Related Work}

The first approaches to learning representations of videos in an unsupervised
way were based on ICA \citep{Hateren, HurriH03}. \citet{QuocISA} approached this
problem using multiple layers of Independent Subspace Analysis modules.  Generative
models for understanding transformations between pairs of consecutive images are
also well studied \citep{memisevic_relational_pami, factoredBoltzmann, morphBM}.
This work was extended recently by \citet{NIPS2014_5549} to model longer
sequences.

Recently, \citet{RanzatoVideo} proposed a generative model for videos. The model
uses a recurrent neural network to predict the next frame or interpolate between
frames. In this work, the authors highlight the importance of choosing the right
loss function.  It is argued that squared loss in input space is not the right
objective because it does not respond well to small distortions in input space.
The proposed solution is to quantize image patches into a large dictionary and
train the model to predict the identity of the target patch. This does solve
some of the problems of squared loss but it introduces an arbitrary dictionary
size into the picture and altogether removes the idea of patches being similar
or dissimilar to one other. 
Designing an appropriate loss
function that respects our notion of visual similarity is a very hard problem
(in a sense, almost as hard as the modeling problem we want to solve in the
first place).  Therefore, in this paper, we use the simple squared loss
objective function as a starting point and focus on designing an encoder-decoder
RNN architecture that can be used with any loss function.

\section{Model Description}

In this section, we describe several variants of our LSTM Encoder-Decoder model.
The basic unit of our network is the LSTM cell block.
Our implementation of LSTMs follows closely the one discussed by \citet{Graves13}.

\subsection{Long Short Term Memory}
\begin{figure}[ht]
\centering
  \includegraphics[clip=true, trim=150 460 160 100,width=0.9\linewidth]{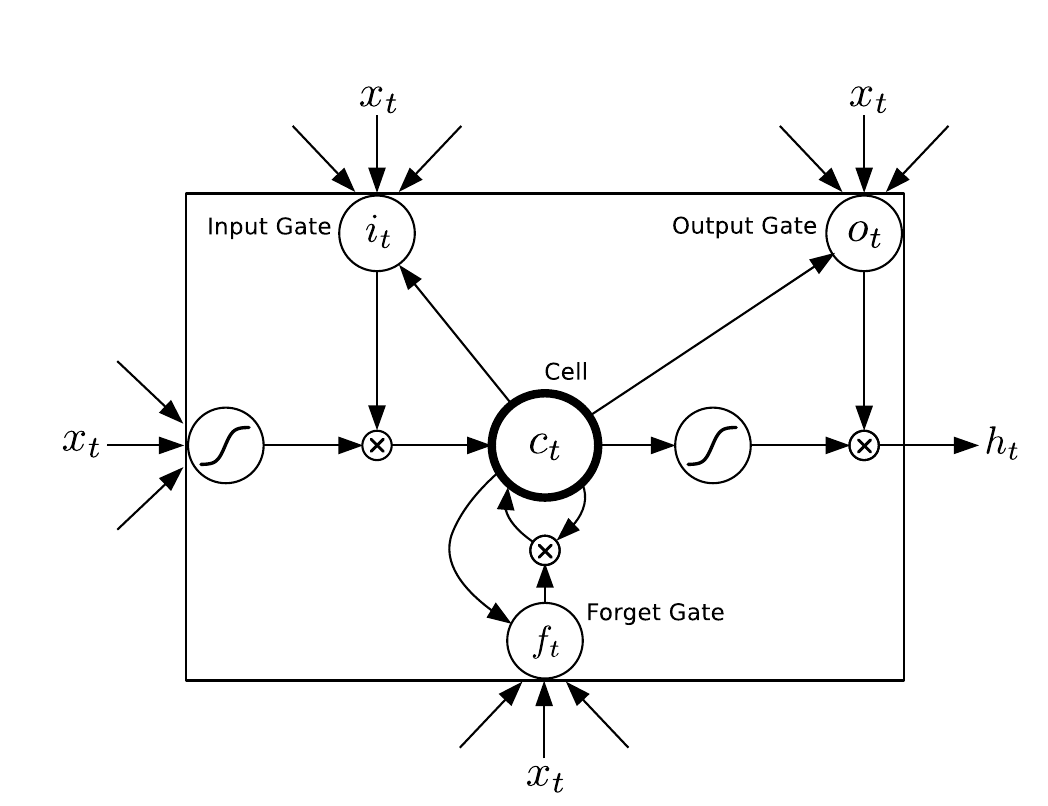}
  \caption{LSTM unit}
  \label{fig:lstm}
  \vspace{-0.2in}
\end{figure}

In this section we briefly describe the LSTM unit which is the basic building block of
our model. The unit is shown in \Figref{fig:lstm} (reproduced from \citet{Graves13}).

Each LSTM unit has a cell which has a state $c_t$ at time $t$. This cell can be
thought of as a memory unit. Access to this memory unit for reading or modifying
it is controlled through sigmoidal gates -- input gate $i_t$, forget gate $f_t$
and output gate $o_t$. The LSTM unit operates as follows.  At each time step it
receives inputs from two external sources at each of the four terminals (the
three gates and the input).  The first source is the current frame ${\xx_t}$.
The second source is the previous hidden states of all LSTM units in the same
layer $\hh_{t-1}$. Additionally, each gate has an internal source, the cell
state $c_{t-1}$ of its cell block. The links between a cell and its own gates
are called \emph{peephole} connections. The inputs coming from different sources
get added up, along with a bias. The gates are activated by passing their total
input through the logistic function. The total input at the input terminal is
passed through the tanh non-linearity. The resulting activation is multiplied by
the activation of the input gate. This is then added to the cell state after
multiplying the cell state by the forget gate's activation $f_t$.  The final output
from the LSTM unit $h_t$ is computed by multiplying the output gate's activation
$o_t$ with the updated cell state passed through a tanh non-linearity. These
updates are summarized for a layer of LSTM units as follows
\vspace{-0.1in}
\begin{eqnarray*}
  \ii_t &=& \sigma\left(W_{xi}\xx_t + W_{hi}\hh_{t-1} + W_{ci}\cc_{t-1} + \bb_i\right),\\
  \ff_t &=& \sigma\left(W_{xf}\xx_t + W_{hf}\hh_{t-1} + W_{cf}\cc_{t-1} + \bb_f\right),\\
  \cc_t &=& \ff_t \cc_{t-1} + \ii_t \tanh\left(W_{xc}\xx_t + W_{hc}\hh_{t-1} + \bb_c\right),\\
  \oo_t &=& \sigma\left(W_{xo}\xx_t + W_{ho}\hh_{t-1} + W_{co}\cc_{t} + \bb_o \right),\\
  \hh_t &=& \oo_t\tanh(\cc_t).
\end{eqnarray*}
  \vspace{-0.3in}

Note that all $W_{c\bullet}$ matrices are diagonal, whereas the rest are dense.
The key advantage of using an LSTM unit over a traditional neuron in an RNN is
that the cell state in an LSTM unit \emph{sums} activities over time. Since derivatives
distribute over sums, the error derivatives don't vanish quickly as they get sent back
into time. This makes it easy to do credit assignment over long sequences and
discover long-range features.

\subsection{LSTM Autoencoder Model}

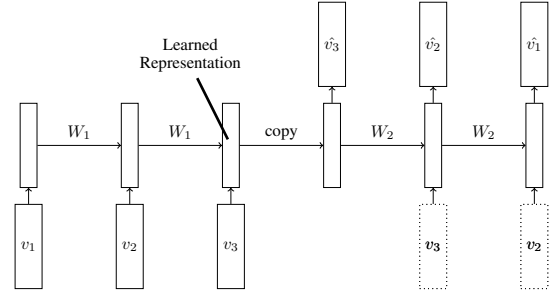
\begin{figure}[t]
  \resizebox{0.9\linewidth}{!}{%
    \makeatletter
\ifx\du\undefined
  \newlength{\du}
\fi
\setlength{\du}{\unitlength}
\ifx\spacing\undefined
  \newlength{\spacing}
\fi
\setlength{\spacing}{60\unitlength}
\begin{tikzpicture}
\pgfsetlinewidth{0.5\du}
\pgfsetmiterjoin
\pgfsetbuttcap

\node[rectangle, draw=black, minimum width=10\du, minimum height=50\du] (v1) at (0\spacing, 0) {$v_1$};
\node[rectangle, draw=black, minimum width=10\du, minimum height=50\du] (v2) at (\spacing, 0) {$v_2$};
\node[rectangle, draw=black, minimum width=10\du, minimum height=50\du] (v3) at (2\spacing, 0) {$v_3$};

\node[rectangle, dotted, draw=black, minimum width=10\du, minimum height=50\du] (v5) at (4\spacing, 0) {$v_3$};
\node[rectangle, dotted, draw=black, minimum width=10\du, minimum height=50\du] (v6) at (5\spacing, 0) {$v_2$};

\node[rectangle, dotted, draw=black, minimum width=10\du, minimum height=50\du] (v5f) at (4\spacing, 0) {$v_3$};
\node[rectangle, dotted, draw=black, minimum width=10\du, minimum height=50\du] (v6f) at (5\spacing, 0) {$v_2$};

\node[rectangle, draw=black, minimum width=10\du, minimum height=50\du] (h1) at (0\spacing, \spacing) {};
\node[rectangle, draw=black, minimum width=10\du, minimum height=50\du] (h2) at (\spacing, \spacing) {};
\node[rectangle, draw=black, minimum width=10\du, minimum height=50\du] (h3) at (2\spacing, \spacing) {};

\node[rectangle, draw=black, minimum width=10\du, minimum height=50\du] (h4) at (3\spacing, \spacing) {};
\node[rectangle, draw=black, minimum width=10\du, minimum height=50\du] (h5) at (4\spacing, \spacing) {};
\node[rectangle, draw=black, minimum width=10\du, minimum height=50\du] (h6) at (5\spacing, \spacing) {};

\node[rectangle, draw=black, minimum width=10\du, minimum height=50\du] (v4r) at (3\spacing, 2\spacing) {$\hat{v_3}$};
\node[rectangle, draw=black, minimum width=10\du, minimum height=50\du] (v5r) at (4\spacing, 2\spacing) {$\hat{v_2}$};
\node[rectangle, draw=black, minimum width=10\du, minimum height=50\du] (v6r) at (5\spacing, 2\spacing) {$\hat{v_1}$};

\node[anchor=center]  at (1.6\spacing, 2\spacing) {Learned};
\node[anchor=center] (p1) at (1.6\spacing, 1.8\spacing) {Representation};
\node[anchor=center] (p2) at (2\spacing, \spacing) {};

\draw[->] (v1) -- (h1);
\draw[->] (v2) -- (h2);
\draw[->] (v3) -- (h3);
\draw[->] (v5) -- (h5);
\draw[->] (v6) -- (h6);

\draw[->] (h1) -- node[above] {$W_1$} (h2);
\draw[->] (h2) -- node[above] {$W_1$} (h3);
\draw[->] (h3) -- node[above] {copy} (h4);
\draw[->] (h4) -- node[above] {$W_2$} (h5);
\draw[->] (h5) -- node[above] {$W_2$} (h6);

\draw[->] (h4) -- (v4r);
\draw[->] (h5) -- (v5r);
\draw[->] (h6) -- (v6r);

\draw[-, ultra thick] (p1) -- (p2);
\end{tikzpicture}
  }
\caption{\small LSTM Autoencoder Model}
\label{fig:autoencoder}
\vspace{-0.1in}
\end{figure}

In this section, we describe a model that uses Recurrent Neural Nets (RNNs) made
of LSTM units to do unsupervised learning.  The model consists of two RNNs --
the encoder LSTM and the decoder LSTM as shown in \Figref{fig:autoencoder}. The
input to the model is a sequence of vectors (image patches or features).  The
encoder LSTM reads in this sequence. After the last input has been read, the
decoder LSTM takes over and outputs a prediction for the target sequence. The
target sequence is same as the input sequence, but in reverse order. Reversing
the target sequence makes the optimization easier because the model can get off
the ground by looking at low range correlations.  This is also inspired by how
lists are represented in LISP.  The encoder can be seen as creating a list by
applying the {\tt cons} function on the previously constructed list and the new
input. The decoder essentially unrolls this list, with the hidden to output
weights extracting the element at the top of the list ({\tt car} function) and
the hidden to hidden weights extracting the rest of the list ({\tt cdr}
function). Therefore, the first element out is the last element in.

The decoder can be of two kinds -- conditional or unconditioned.
A conditional decoder receives the last generated output frame as
input, i.e., the dotted input in \Figref{fig:autoencoder} is present.
An unconditioned decoder does not receive that input. This is discussed in more
detail in \Secref{sec:condvsuncond}. \Figref{fig:autoencoder} shows a single
layer LSTM Autoencoder. The architecture can be extend to multiple layers by
stacking LSTMs on top of each other.

\emph{Why should this learn good features?}\\
The state of the encoder LSTM after the last input has been read is the
representation of the input video. The decoder LSTM is being asked to
reconstruct back the input sequence from this representation. In order to do so,
the representation must retain information about the appearance of the objects
and the background as well as the motion contained in the video. 
However, an important question for any autoencoder-style model is what prevents
it from learning an identity mapping and effectively copying the input to the
output. In that case all the information about the input would still be present
but the representation will be no better than the input.  There are two factors
that control this behaviour. First, the fact that there are only a fixed number
of hidden units makes it unlikely that the model can learn trivial mappings for
arbitrary length input sequences. Second, the same LSTM operation is used to
decode the representation recursively. This means that the same dynamics must be
applied on the representation at any stage of decoding. This further prevents
the model from learning an identity mapping.

\subsection{LSTM Future Predictor Model}

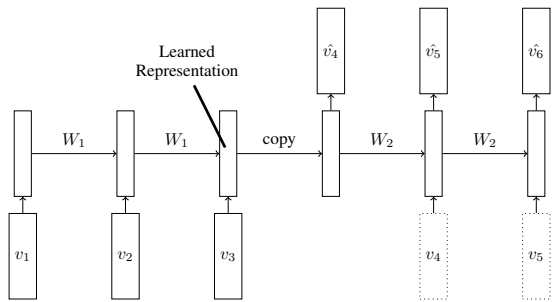
\begin{figure}[t]
  \resizebox{0.9\linewidth}{!}{%
    \makeatletter
\ifx\du\undefined
  \newlength{\du}
\fi
\setlength{\du}{\unitlength}
\ifx\spacing\undefined
  \newlength{\spacing}
\fi
\setlength{\spacing}{60\unitlength}
\begin{tikzpicture}
\pgfsetlinewidth{0.5\du}
\pgfsetmiterjoin
\pgfsetbuttcap

\node[rectangle, draw=black, minimum width=10\du, minimum height=50\du] (v1) at (0\spacing, 0) {$v_1$};
\node[rectangle, draw=black, minimum width=10\du, minimum height=50\du] (v2) at (\spacing, 0) {$v_2$};
\node[rectangle, draw=black, minimum width=10\du, minimum height=50\du] (v3) at (2\spacing, 0) {$v_3$};

\node[rectangle, dotted, draw=black, minimum width=10\du, minimum height=50\du] (v5) at (4\spacing, 0) {$v_4$};
\node[rectangle, dotted, draw=black, minimum width=10\du, minimum height=50\du] (v6) at (5\spacing, 0) {$v_5$};

\node[rectangle, draw=black, minimum width=10\du, minimum height=50\du] (h1) at (0\spacing, \spacing) {};
\node[rectangle, draw=black, minimum width=10\du, minimum height=50\du] (h2) at (\spacing, \spacing) {};
\node[rectangle, draw=black, minimum width=10\du, minimum height=50\du] (h3) at (2\spacing, \spacing) {};

\node[rectangle, draw=black, minimum width=10\du, minimum height=50\du] (h4) at (3\spacing, \spacing) {};
\node[rectangle, draw=black, minimum width=10\du, minimum height=50\du] (h5) at (4\spacing, \spacing) {};
\node[rectangle, draw=black, minimum width=10\du, minimum height=50\du] (h6) at (5\spacing, \spacing) {};

\node[rectangle, draw=black, minimum width=10\du, minimum height=50\du] (v4r) at (3\spacing, 2\spacing) {$\hat{v_4}$};
\node[rectangle, draw=black, minimum width=10\du, minimum height=50\du] (v5r) at (4\spacing, 2\spacing) {$\hat{v_5}$};
\node[rectangle, draw=black, minimum width=10\du, minimum height=50\du] (v6r) at (5\spacing, 2\spacing) {$\hat{v_6}$};

\node[anchor=center]  at (1.6\spacing, 2\spacing) {Learned};
\node[anchor=center] (p1) at (1.6\spacing, 1.8\spacing) {Representation};
\node[anchor=center] (p2) at (2\spacing, \spacing) {};

\draw[->] (v1) -- (h1);
\draw[->] (v2) -- (h2);
\draw[->] (v3) -- (h3);

\draw[->] (v5) -- (h5);
\draw[->] (v6) -- (h6);

\draw[->] (h1) -- node[above] {$W_1$} (h2);
\draw[->] (h2) -- node[above] {$W_1$} (h3);
\draw[->] (h3) -- node[above] {copy} (h4);
\draw[->] (h4) -- node[above] {$W_2$} (h5);
\draw[->] (h5) -- node[above] {$W_2$} (h6);

\draw[->] (h4) -- (v4r);
\draw[->] (h5) -- (v5r);
\draw[->] (h6) -- (v6r);
\draw[-, ultra thick] (p1) -- (p2);
\end{tikzpicture}
  }
\caption{\small LSTM Future Predictor Model}
\label{fig:futurepredictor}
\vspace{-0.1in}
\end{figure}

Another natural unsupervised learning task for sequences is predicting the
future.  This is the approach used in language models for modeling sequences of
words. The design of the Future Predictor Model is same as that of the
Autoencoder Model, except that the decoder LSTM in this case predicts frames of
the video that come after the input sequence (\Figref{fig:futurepredictor}).
\citet{RanzatoVideo} use a similar model but predict only the next frame at each
time step. This model, on the other hand, predicts a long sequence into the
future. Here again we can consider two variants of the decoder -- conditional
and unconditioned.

\emph{Why should this learn good features?}\\
In order to predict the next few frames correctly, the model needs information
about which objects and background are present and how they are moving so that
the motion can be extrapolated. The hidden state coming out from the
encoder will try to capture this information. Therefore, this state can be seen as
a representation of the input sequence.

\subsection{Conditional Decoder}
\label{sec:condvsuncond}

For each of these two models, we can consider two possibilities - one in which
the decoder LSTM is conditioned on the last generated frame and the other in
which it is not.  In the experimental section, we explore these choices
quantitatively. Here we briefly discuss arguments for and against a conditional
decoder.  A strong argument in favour of using a conditional decoder is that it
allows the decoder to model multiple modes in the target sequence distribution.
Without that, we would end up averaging the multiple modes in the low-level
input space.  However, this is an issue only if we expect multiple modes in the
target sequence distribution. For the LSTM Autoencoder, there is only one
correct target and hence a unimodal target distribution. But for the LSTM Future
Predictor there is a possibility of multiple targets given an input because even
if we assume a deterministic universe, everything needed to predict the future
will not necessarily be observed in the input.

There is also an argument against using a conditional decoder from the
optimization point-of-view. There are strong short-range correlations in
video data, for example, most of the content of a frame is same as the previous
one. If the decoder was given access to the last few frames while generating a
particular frame at training time, it would find it easy to pick up on these
correlations. There would only be a very small gradient that tries to fix up the
extremely subtle errors that require long term knowledge about the input
sequence. In an unconditioned decoder, this input is removed and the model is
forced to look for information deep inside the encoder.

\subsection{A Composite Model}
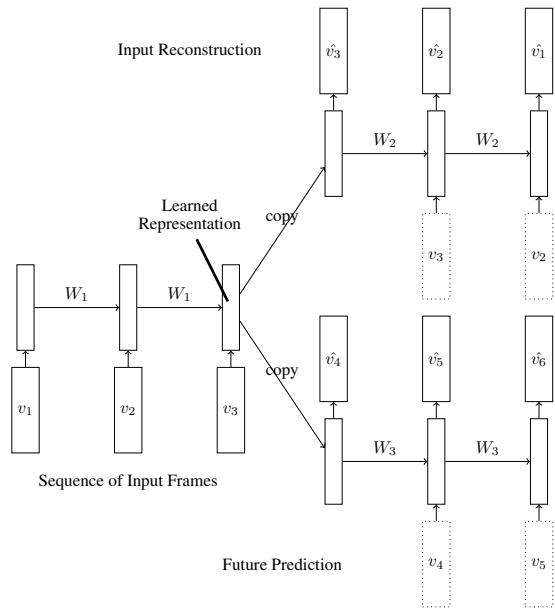
\begin{figure}[t]
\resizebox{0.9\linewidth}{!}{%
  \makeatletter
\ifx\du\undefined
  \newlength{\du}
\fi
\setlength{\du}{\unitlength}
\ifx\spacing\undefined
  \newlength{\spacing}
\fi
\setlength{\spacing}{60\unitlength}
\begin{tikzpicture}
\pgfsetlinewidth{0.5\du}
\pgfsetmiterjoin
\pgfsetbuttcap

\node[rectangle, draw=black, minimum width=10\du, minimum height=50\du] (v1) at (0\spacing, 1.5\spacing) {$v_1$};
\node[rectangle, draw=black, minimum width=10\du, minimum height=50\du] (v2) at (\spacing, 1.5\spacing) {$v_2$};
\node[rectangle, draw=black, minimum width=10\du, minimum height=50\du] (v3) at (2\spacing, 1.5\spacing) {$v_3$};

\node[rectangle, dotted, draw=black, minimum width=10\du, minimum height=50\du] (v5) at (4\spacing, 3\spacing) {$v_3$};
\node[rectangle, dotted, draw=black, minimum width=10\du, minimum height=50\du] (v6) at (5\spacing, 3\spacing) {$v_2$};

\node[rectangle, dotted, draw=black, minimum width=10\du, minimum height=50\du] (v5f) at (4\spacing, 0) {$v_4$};
\node[rectangle, dotted, draw=black, minimum width=10\du, minimum height=50\du] (v6f) at (5\spacing, 0) {$v_5$};

\node[rectangle, draw=black, minimum width=10\du, minimum height=50\du] (h1) at (0\spacing, 2.5\spacing) {};
\node[rectangle, draw=black, minimum width=10\du, minimum height=50\du] (h2) at (\spacing, 2.5\spacing) {};
\node[rectangle, draw=black, minimum width=10\du, minimum height=50\du] (h3) at (2\spacing, 2.5\spacing) {};

\node[rectangle, draw=black, minimum width=10\du, minimum height=50\du] (h4) at (3\spacing, 4\spacing) {};
\node[rectangle, draw=black, minimum width=10\du, minimum height=50\du] (h5) at (4\spacing, 4\spacing) {};
\node[rectangle, draw=black, minimum width=10\du, minimum height=50\du] (h6) at (5\spacing, 4\spacing) {};

\node[rectangle, draw=black, minimum width=10\du, minimum height=50\du] (h4f) at (3\spacing, \spacing) {};
\node[rectangle, draw=black, minimum width=10\du, minimum height=50\du] (h5f) at (4\spacing, \spacing) {};
\node[rectangle, draw=black, minimum width=10\du, minimum height=50\du] (h6f) at (5\spacing, \spacing) {};

\node[rectangle, draw=black, minimum width=10\du, minimum height=50\du] (v4r) at (3\spacing, 5\spacing) {$\hat{v_3}$};
\node[rectangle, draw=black, minimum width=10\du, minimum height=50\du] (v5r) at (4\spacing, 5\spacing) {$\hat{v_2}$};
\node[rectangle, draw=black, minimum width=10\du, minimum height=50\du] (v6r) at (5\spacing, 5\spacing) {$\hat{v_1}$};

\node[rectangle, draw=black, minimum width=10\du, minimum height=50\du] (v4rf) at (3\spacing, 2\spacing) {$\hat{v_4}$};
\node[rectangle, draw=black, minimum width=10\du, minimum height=50\du] (v5rf) at (4\spacing, 2\spacing) {$\hat{v_5}$};
\node[rectangle, draw=black, minimum width=10\du, minimum height=50\du] (v6rf) at (5\spacing, 2\spacing) {$\hat{v_6}$};

\node[anchor=center]  at (\spacing, 0.8\spacing) {Sequence of Input Frames};
\node[anchor=center]  at (2.5\spacing, 0\spacing) {Future Prediction};
\node[anchor=center]  at (1.6\spacing, 5\spacing) {Input Reconstruction};
\node[anchor=center]  at (1.6\spacing, 3.5\spacing) {Learned};
\node[anchor=center] (p1) at (1.6\spacing, 3.3\spacing) {Representation};
\node[anchor=center] (p2) at (2\spacing, 2.5\spacing) {};

\draw[-, ultra thick] (p1) -- (p2);

\draw[->] (v1) -- (h1);
\draw[->] (v2) -- (h2);
\draw[->] (v3) -- (h3);
\draw[->] (v5) -- (h5);
\draw[->] (v6) -- (h6);
\draw[->] (v5f) -- (h5f);
\draw[->] (v6f) -- (h6f);

\draw[->] (h1) -- node[above] {$W_1$} (h2);
\draw[->] (h2) -- node[above] {$W_1$} (h3);
\draw[->] (h3) -- node[above] {copy} (h4);
\draw[->] (h3) -- node[above] {copy} (h4f);
\draw[->] (h4) -- node[above] {$W_2$} (h5);
\draw[->] (h5) -- node[above] {$W_2$} (h6);
\draw[->] (h4f) -- node[above] {$W_3$} (h5f);
\draw[->] (h5f) -- node[above] {$W_3$} (h6f);

\draw[->] (h4) -- (v4r);
\draw[->] (h5) -- (v5r);
\draw[->] (h6) -- (v6r);
\draw[->] (h4f) -- (v4rf);
\draw[->] (h5f) -- (v5rf);
\draw[->] (h6f) -- (v6rf);
\end{tikzpicture}
}
\caption{\small The Composite Model: The LSTM predicts the future as well as the input sequence.}
\label{fig:modelcombo}
\vspace{-0.2in}
\end{figure}

The two tasks -- reconstructing the input and predicting the future can be
combined to create a composite model as shown in \Figref{fig:modelcombo}. Here
the encoder LSTM is asked to come up with a state from which we can \emph{both} predict
the next few frames as well as reconstruct the input.

This composite model tries to overcome the shortcomings that each model suffers
on its own. A high-capacity autoencoder would suffer from the tendency to learn
trivial representations that just memorize the inputs. However, this
memorization is not useful at all for predicting the future. Therefore, the
composite model cannot just memorize information. On the other hand, the future
predictor suffers form the tendency to store information only about the last few
frames since those are most important for predicting the future, i.e., in order
to predict $v_{t}$, the frames $\{v_{t-1}, \ldots, v_{t-k}\}$ are much more
important than $v_0$, for some small value of $k$. Therefore the representation
at the end of the encoder will have forgotten about a large part of the input.
But if we ask the model to also predict \emph {all} of the input sequence, then
it cannot just pay attention to the last few frames.

\section{Experiments}
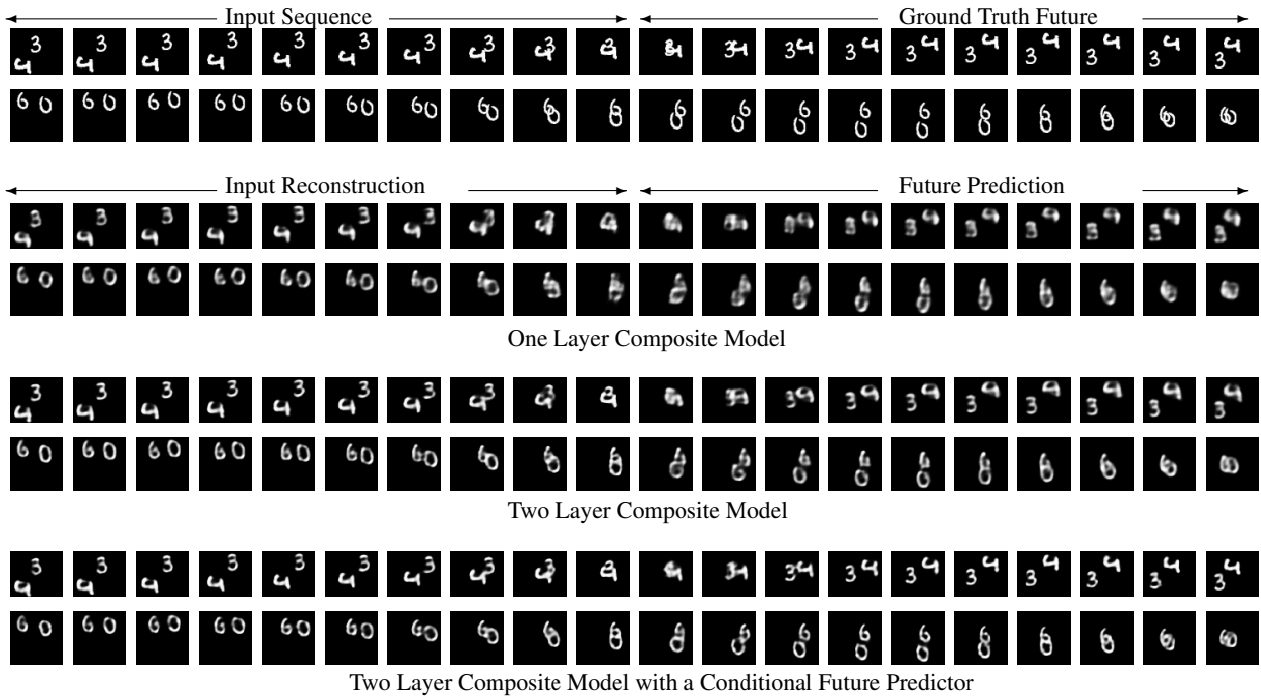
\begin{figure*}[ht]
\centering

\includegraphics[width=\textwidth]{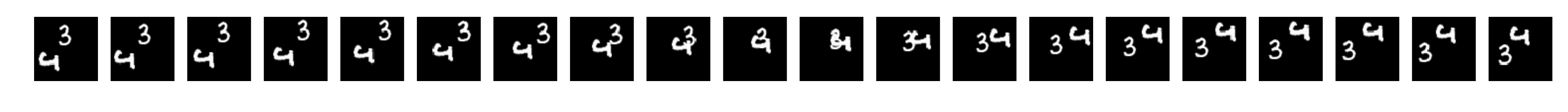}

\vspace{-0.18in}
\includegraphics[width=\textwidth]{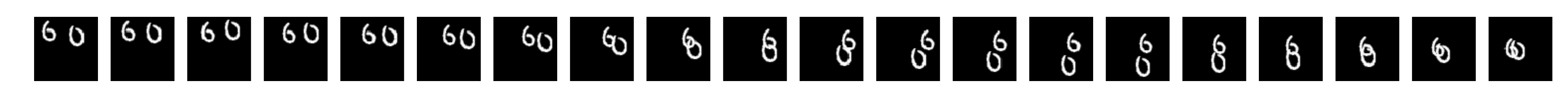}

\vspace{0.1in}
\includegraphics[width=\textwidth]{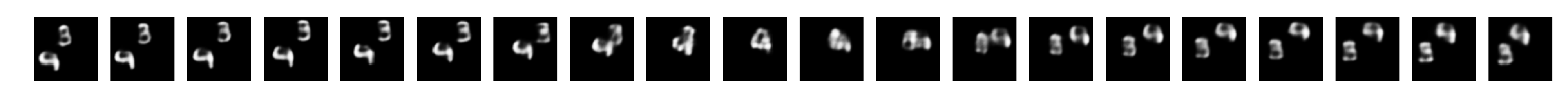}

\vspace{-0.18in}
\includegraphics[width=\textwidth]{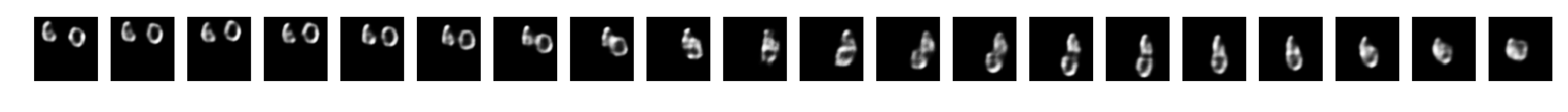}

\vspace{0.1in}
\includegraphics[width=\textwidth]{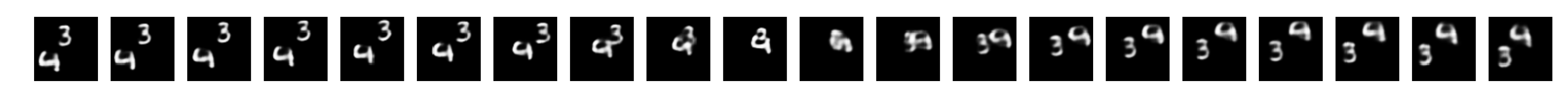}

\vspace{-0.18in}
\includegraphics[width=\textwidth]{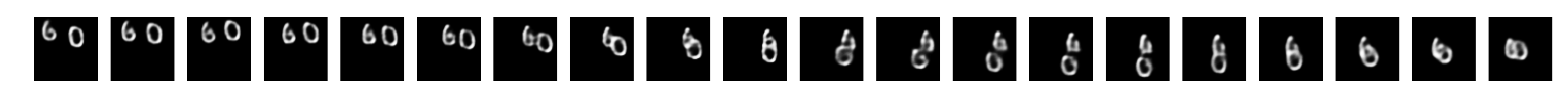}

\vspace{0.1in}
\includegraphics[width=\textwidth]{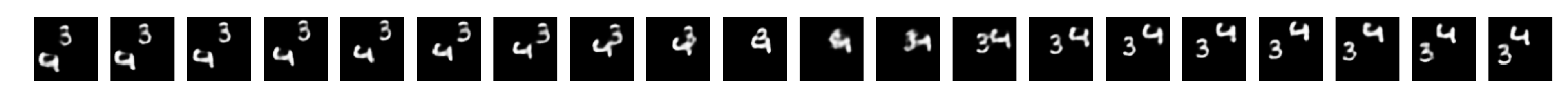}

\vspace{-0.18in}
\includegraphics[width=\textwidth]{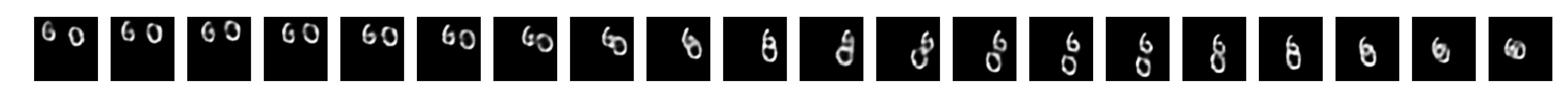}
\small
\begin{picture}(0, 0)(-245, -200)
\put(-142, 62){Ground Truth Future}
\put(-50, 64) {\vector(1, 0){40}}
\put(-150, 64) {\vector(-1, 0){90}}
\put(-397, 62){Input Sequence}
\put(-400, 64) {\vector(-1, 0){80}}
\put(-335, 64) {\vector(1, 0){90}}

\put(-142, -2){Future Prediction}
\put(-50, -1) {\vector(1, 0){40}}
\put(-150, -1) {\vector(-1, 0){90}}
\put(-397, -2){Input Reconstruction}
\put(-400, -1) {\vector(-1, 0){80}}
\put(-305, -1) {\vector(1, 0){60}}

\put(-290, -60){One Layer Composite Model}
\put(-290, -125){Two Layer Composite Model}
\put(-350, -190){Two Layer Composite Model with a Conditional Future Predictor}

\end{picture}
\caption{\small Reconstruction and future prediction obtained from the Composite Model
on a dataset of moving MNIST digits.}
\label{fig:bouncing_mnist}
\vspace{-0.2in}
\end{figure*}

We design experiments to accomplish the following objectives:
\begin{itemize}
\vspace{-0.1in}
\item Get a qualitative understanding of what the LSTM learns to do.
\item Measure the benefit of initializing networks for supervised learning tasks
with the weights found by unsupervised learning, especially with very few training examples.
\item Compare the different proposed models - Autoencoder, Future Predictor and
Composite models and their conditional variants.
\item Compare with state-of-the-art action recognition benchmarks.
\end{itemize}

\subsection{Datasets}
We use the UCF-101 and HMDB-51 datasets for supervised tasks.
The UCF-101 dataset \citep{UCF101} contains 13,320 videos with an average length of
6.2 seconds belonging to 101 different action categories. The dataset has 3
standard train/test splits with the training set containing around 9,500 videos
in each split (the rest are test).
The HMDB-51 dataset \citep{HMDB} contains 5100 videos belonging to 51 different
action categories. Mean length of the videos is 3.2 seconds. This also has 3
train/test splits with 3570 videos in the training set and rest in test.

To train the unsupervised models, we used a subset of the Sports-1M dataset
\citep{KarpathyCVPR14}, that contains 1~million YouTube clips. 
 Even though this dataset is labelled for actions, we did
not do any supervised experiments on it because of logistical constraints with
working with such a huge dataset. We instead collected 300 hours of video by
randomly sampling 10 second clips from the dataset. It is possible to collect
better samples if instead of choosing randomly, we extracted videos where a lot of
motion is happening and where there are no shot boundaries. However, we did not
do so in the spirit of unsupervised learning, and because we did not want to
introduce any unnatural bias in the samples. We also used the supervised
datasets (UCF-101 and HMDB-51) for unsupervised training. However, we found that
using them did not give any significant advantage over just using the YouTube
videos.

We extracted percepts using the convolutional neural net model of
\citet{Simonyan14c}. The videos have a resolution of 240 $\times$ 320 and were
sampled at almost 30 frames per second. We took the central 224 $\times$ 224
patch from each frame and ran it through the convnet. This gave us the RGB
percepts. Additionally, for UCF-101, we computed flow percepts by extracting flows using the Brox
method and training the temporal stream convolutional network as described by
\citet{Simonyan14b}. We found that the fc6 features worked better than fc7 for
single frame classification using both RGB and flow percepts.  Therefore, we
used the 4096-dimensional fc6 layer as the input representation of our data.
Besides these percepts, we also trained the proposed models on 32 $\times$ 32
patches of pixels.

All models were trained using backprop on a single NVIDIA Titan GPU. A two layer
2048 unit Composite model that predicts 13 frames and reconstructs 16 frames
took 18-20 hours to converge on 300 hours of percepts. We initialized weights
by sampling from a uniform distribution whose scale was set to 1/sqrt(fan-in).
Biases at all the gates were initialized to zero. Peep-hole connections were
initialized to zero. The supervised classifiers trained on 16 frames took 5-15
minutes to converge. The code can be found at \url{https://github.com/emansim/unsupervised-videos}.

\subsection{Visualization and Qualitative Analysis}

\begin{figure*}[ht]
\centering

\includegraphics[clip=true, trim=252 0 0 0, width=\textwidth]{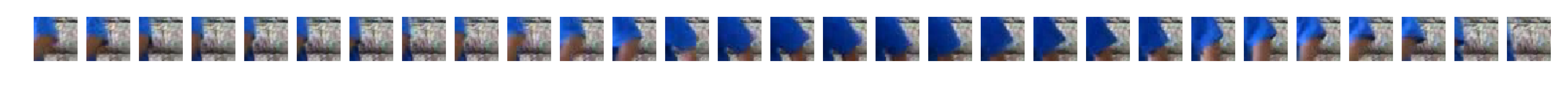}

\vspace{-0.18in}
\includegraphics[clip=true, trim=252 0 0 0, width=\textwidth]{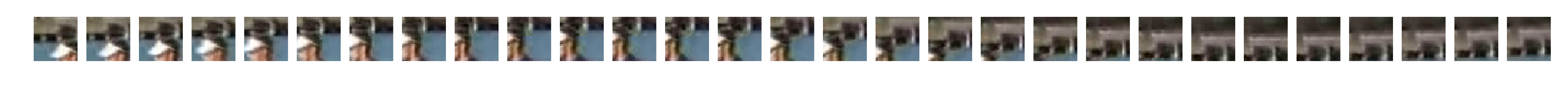}

\vspace{0.1in}
\includegraphics[clip=true, trim=252 0 0 0, width=\textwidth]{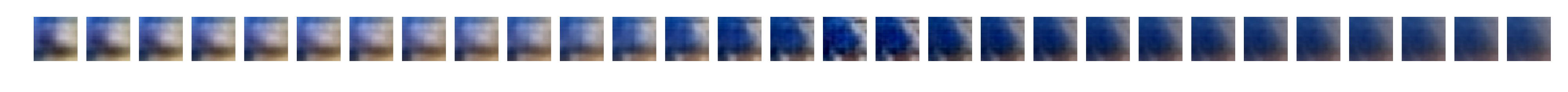}

\vspace{-0.18in}
\includegraphics[clip=true, trim=252 0 0 0, width=\textwidth]{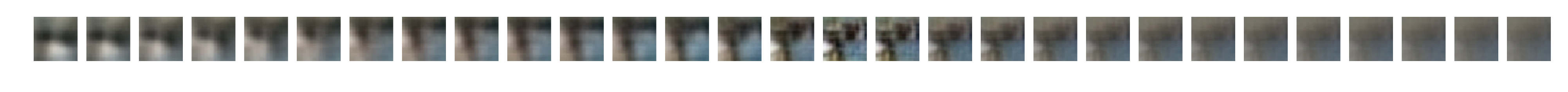}

\vspace{0.1in}
\includegraphics[clip=true, trim=252 0 0 0, width=\textwidth]{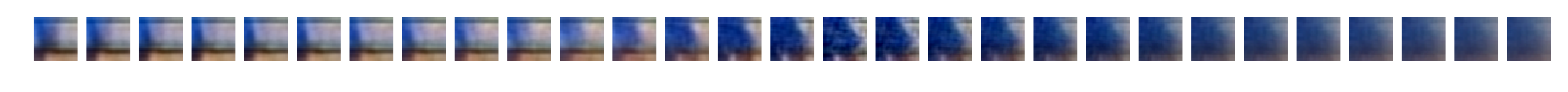}

\vspace{-0.18in}
\includegraphics[clip=true, trim=252 0 0 0, width=\textwidth]{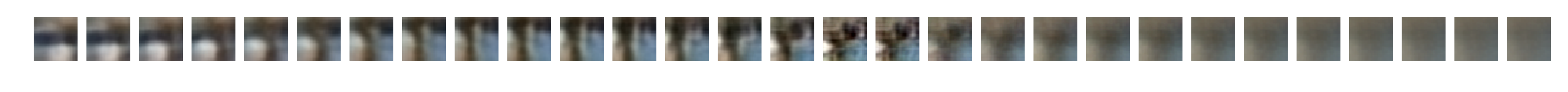}

\small
\begin{picture}(0, 0)(-245, -150)
\put(-142, 60){Ground Truth Future}
\put(-50, 62) {\vector(1, 0){40}}
\put(-150, 62) {\vector(-1, 0){125}}
\put(-397, 60){Input Sequence}
\put(-400, 62) {\vector(-1, 0){80}}
\put(-335, 62) {\vector(1, 0){58}}

\put(-142, -10){Future Prediction}
\put(-50, -9) {\vector(1, 0){40}}
\put(-150, -9) {\vector(-1, 0){125}}
\put(-397, -10){Input Reconstruction}
\put(-400, -9) {\vector(-1, 0){80}}
\put(-320, -9) {\vector(1, 0){43}}

\put(-340, -67){Two Layer Composite Model with 2048 LSTM units}
\put(-340, -135){Two Layer Composite Model with 4096 LSTM units}

\end{picture}
\vspace{-0.2in}
\caption{\small Reconstruction and future prediction obtained from the Composite Model
on a dataset of natural image patches.
The first two rows show ground truth sequences. The model takes 16 frames as inputs.
Only the last 10 frames of the input sequence are shown here. The next 13 frames are the ground truth future. In the rows that
follow, we show the reconstructed and predicted frames for two instances of
the model.
}
\label{fig:image_patches}
\vspace{-0.2in}
\end{figure*}

The aim of this set of experiments to visualize the properties of the proposed
models.

\emph{Experiments on MNIST} \\
We first trained our models on a dataset of moving MNIST digits. In this
dataset, each
video was 20 frames long and consisted of two digits moving inside a 64 $\times$ 64
patch. The digits were chosen randomly from the training set and placed
initially at random locations inside the patch. Each digit was assigned a
velocity whose direction was chosen uniformly randomly on a unit circle and
whose magnitude was also chosen uniformly at random over a fixed range. The
digits bounced-off the edges of the 64 $\times$ 64 frame and overlapped if they
were at the same location. The reason for working with this dataset is that it is
infinite in size and can be generated quickly on the fly. This makes it possible
to explore the model without expensive disk accesses or overfitting issues. It
also has interesting behaviours due to occlusions and the dynamics of bouncing
off the walls.

We first trained a single layer Composite Model. Each LSTM had 2048 units.  The
encoder took 10 frames as input. The decoder tried to reconstruct these 10
frames and the future predictor attempted to predict the next 10 frames. We
used logistic output units with a cross entropy loss function.
\Figref{fig:bouncing_mnist} shows two examples of running this model. The true
sequences are shown in the first two rows. The next two rows show the
reconstruction and future prediction from the one layer Composite Model. It is
interesting to note that the model figures out how to separate superimposed
digits and can model them even as they pass through each other. This shows some
evidence of \emph{disentangling} the two independent factors of variation in
this sequence. The model can also correctly predict the motion after bouncing
off the walls. In order to see if adding depth helps, we trained a two layer
Composite Model, with each layer having 2048 units. We can see that adding
depth helps the model make better predictions. Next, we changed the future
predictor by making it conditional. We can see that this model makes sharper
predictions.

\emph{Experiments on Natural Image Patches} \\
Next, we tried to see if our models can also work with natural image patches.
For this, we trained the models on sequences of 32 $\times$ 32 natural image
patches extracted from the UCF-101 dataset. In this case, we used linear output
units and the squared error loss function. The input was 16 frames and the model
was asked to reconstruct the 16 frames and predict the future 13 frames.
\Figref{fig:image_patches} shows the results obtained from a two layer Composite
model with 2048 units. We found that the reconstructions and the predictions are
both very blurry.  We then trained a bigger model with 4096 units. The outputs
from this model are also shown in \Figref{fig:image_patches}. We can see that
the reconstructions get much sharper.

\begin{figure*}[ht]
  \centering
  \subfigure[Trained Future Predictor]{
\label{fig:gatepatterntrained}
  \includegraphics[clip=true, trim=150 50 120 50,width=0.9\textwidth]{gate_pattern.pdf}}
  \subfigure[Randomly Initialized Future Predictor]{
\label{fig:gatepatternrandom}
  \includegraphics[clip=true, trim=150 50 120 50,width=0.9\textwidth]{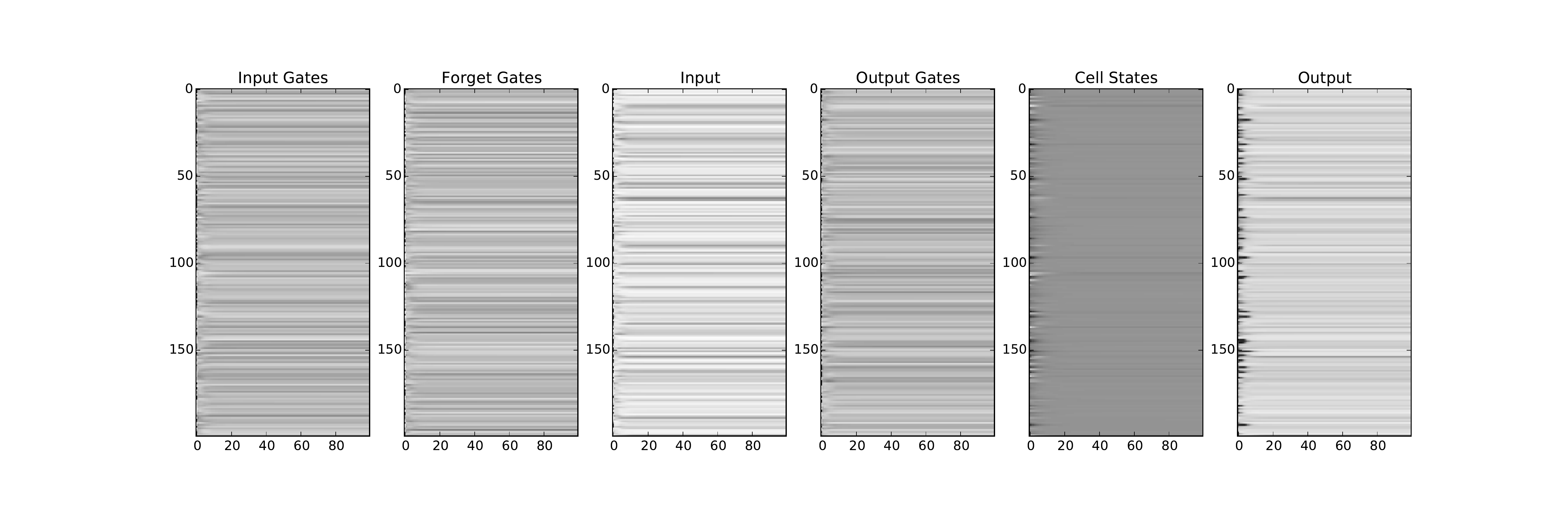}}
\vspace{-0.1in}
\caption{Pattern of activity in 200 randomly chosen LSTM units in the Future
  Predictor of a 1 layer (unconditioned) Composite Model trained on moving MNIST digits.
The vertical axis corresponds to different LSTM units. The horizontal axis is
time. The model was only trained to predict the next 10 frames, but here we let it run to predict
the next 100 frames.
{\bf Top}: The dynamics has a periodic quality which does not die
out. {\bf Bottom} : The pattern of activity, if the trained weights
in the future predictor are replaced by random weights. The dynamics quickly
dies out.
}
\label{fig:gatepattern}
\vspace{-0.2in}
\end{figure*}

\begin{figure*}
\centering

\includegraphics[width=0.9\textwidth]{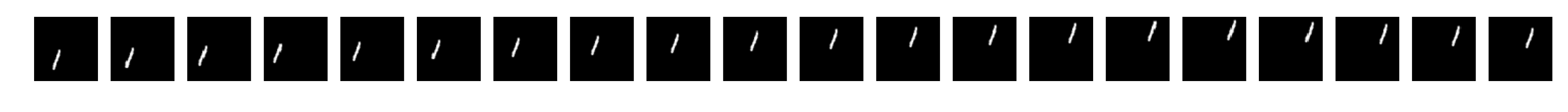}

\vspace{-0.18in}
\includegraphics[width=0.9\textwidth]{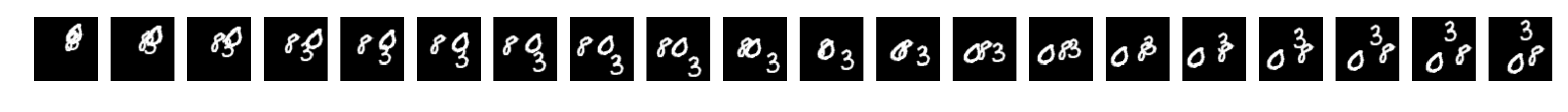}

\vspace{0.1in}
\includegraphics[width=0.9\textwidth]{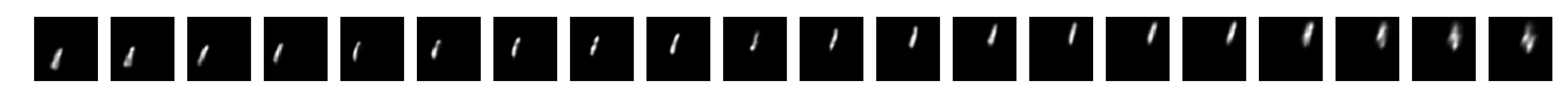}

\vspace{-0.18in}
\includegraphics[width=0.9\textwidth]{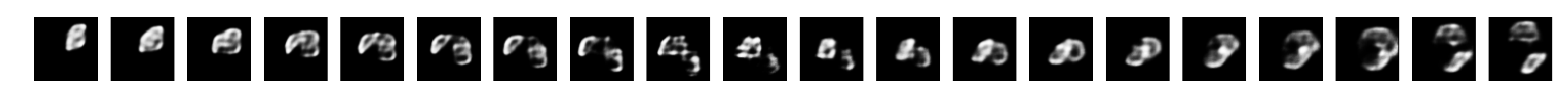}
\vspace{-0.2in}

\small
\begin{picture}(0, 0)(-245, -45)
\put(-172, 62){Ground Truth Future}
\put(-90, 64) {\vector(1, 0){60}}
\put(-180, 64) {\vector(-1, 0){60}}
\put(-397, 62){Input Sequence}
\put(-400, 64) {\vector(-1, 0){55}}
\put(-335, 64) {\vector(1, 0){90}}

\put(-150, 0){Future Prediction}
\put(-80, 1) {\vector(1, 0){50}}
\put(-155, 1) {\vector(-1, 0){85}}
\put(-390, 0){Input Reconstruction}
\put(-395, 1) {\vector(-1, 0){60}}
\put(-305, 1) {\vector(1, 0){60}}
\end{picture}
\caption{\small Out-of-domain runs. Reconstruction and Future prediction for
test sequences of one and three moving digits. The model was trained on
sequences of two moving digits.}
\label{fig:out_of_domain}
\vspace{-0.18in}
\end{figure*}

\emph{Generalization over time scales} \\
In the next experiment, we test if the model can work at time scales that are
different than what it was trained on.  We take a one hidden layer unconditioned
Composite Model trained on moving MNIST digits. The model has 2048 LSTM units
and looks at a 64 $\times$ 64 input. It was trained on input sequences of 10
frames to reconstruct those 10 frames as well as predict 10 frames into the
future. In order to test if the future predictor is able to generalize beyond 10
frames, we let the model run for 100 steps into the future.
\Figref{fig:gatepatterntrained} shows the pattern of activity in the LSTM units of the
future predictor
pathway for a randomly chosen test input. It shows the activity at each of the
three sigmoidal gates (input, forget, output), the input (after the tanh
non-linearity, before being multiplied by the input gate), the cell state and
the final output (after being multiplied by the output gate).  Even though the
units are ordered randomly along the vertical axis, we can see that the dynamics
has a periodic quality to it. The model is able to generate persistent motion
for long periods of time. In terms of reconstruction, the model only outputs
blobs after the first 15 frames, but the motion is relatively well preserved.
More results, including long range future predictions over hundreds of time steps can see been at
\url{http://www.cs.toronto.edu/~nitish/unsupervised_video}.
To show that setting up a periodic behaviour is not trivial,
\Figref{fig:gatepatternrandom} shows the activity from a randomly initialized future
predictor. Here, the LSTM state quickly converges and the outputs blur completely.

\emph{Out-of-domain Inputs} \\
Next, we test this model's ability to deal with out-of-domain inputs. For this,
we test the model on sequences of one and three moving digits. The model was
trained on sequences of two moving digits, so it has never seen inputs with just
one digit or three digits. \Figref{fig:out_of_domain} shows the reconstruction
and future prediction results. For one moving digit, we can see that the model
can do a good job but it really tries to hallucinate a second digit overlapping
with the first one. The second digit shows up towards the end of the future
reconstruction.  For three digits, the model merges digits into blobs. However,
it does well at getting the overall motion right. This highlights a key drawback
of modeling entire frames of input in a single pass.  In order to model videos
with variable number of objects, we perhaps need models that not only have an attention
mechanism in place, but can also learn to execute themselves a variable number
of times and do variable amounts of computation.

\begin{figure*}
  \centering
  \small
  \subfigure[\small Inputs]{
  \includegraphics[clip=true, trim=0 0 0 100,width=0.48\textwidth]{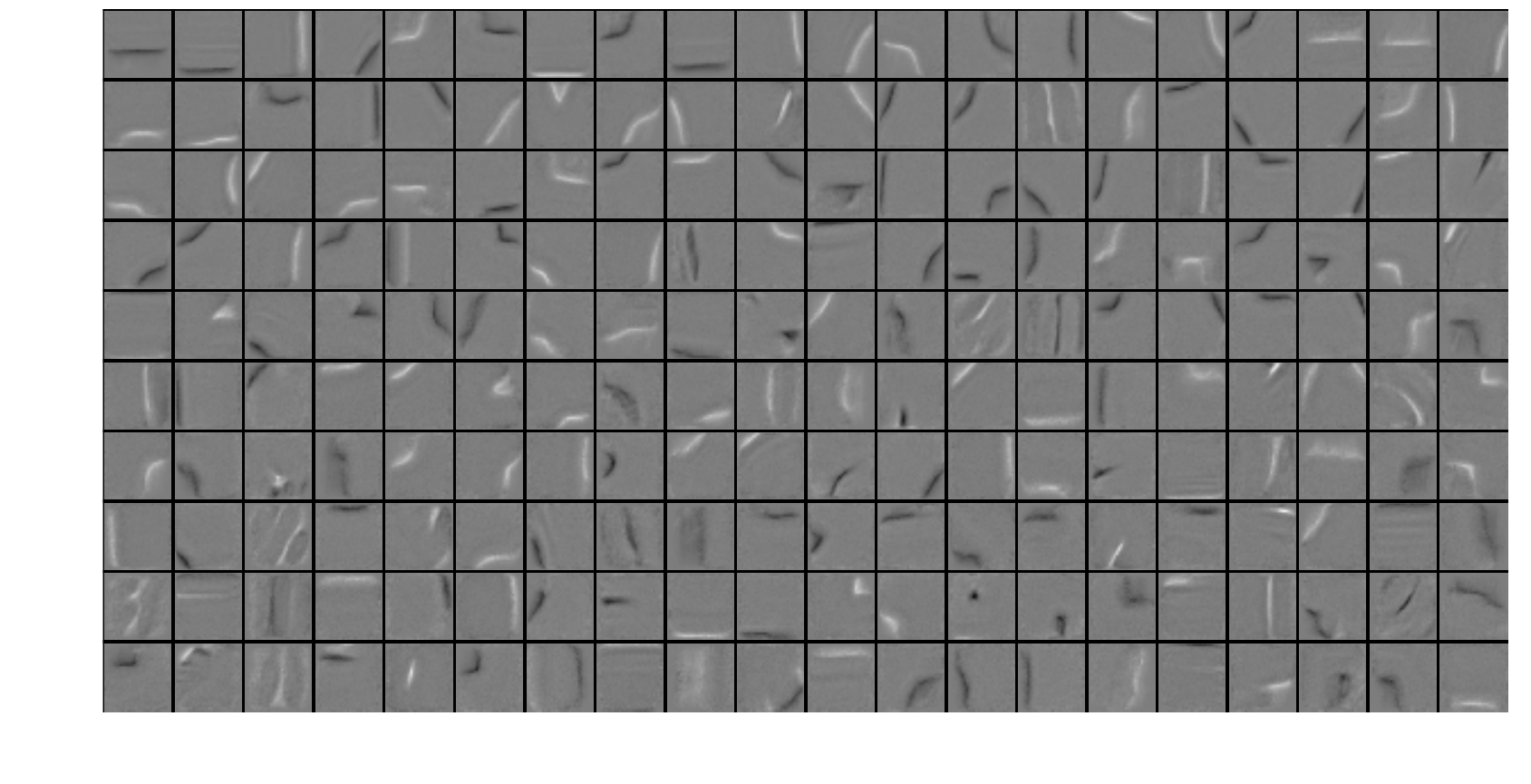}}
  \subfigure[\small Input Gates]{
  \includegraphics[clip=true, trim=0 0 0 100,width=0.48\textwidth]{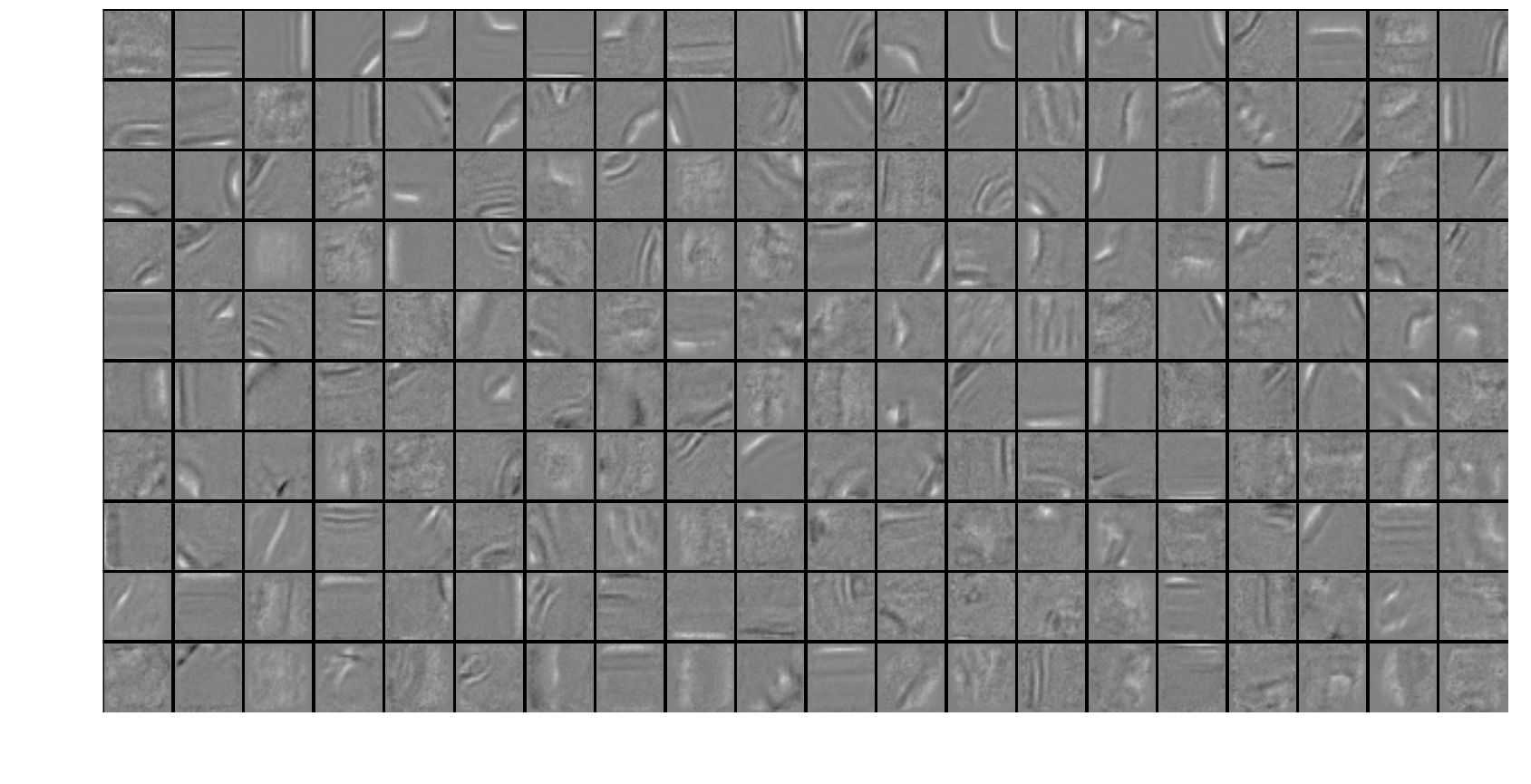}}
  \subfigure[\small Forget Gates]{
  \includegraphics[clip=true, trim=0 0 0 100,width=0.48\textwidth]{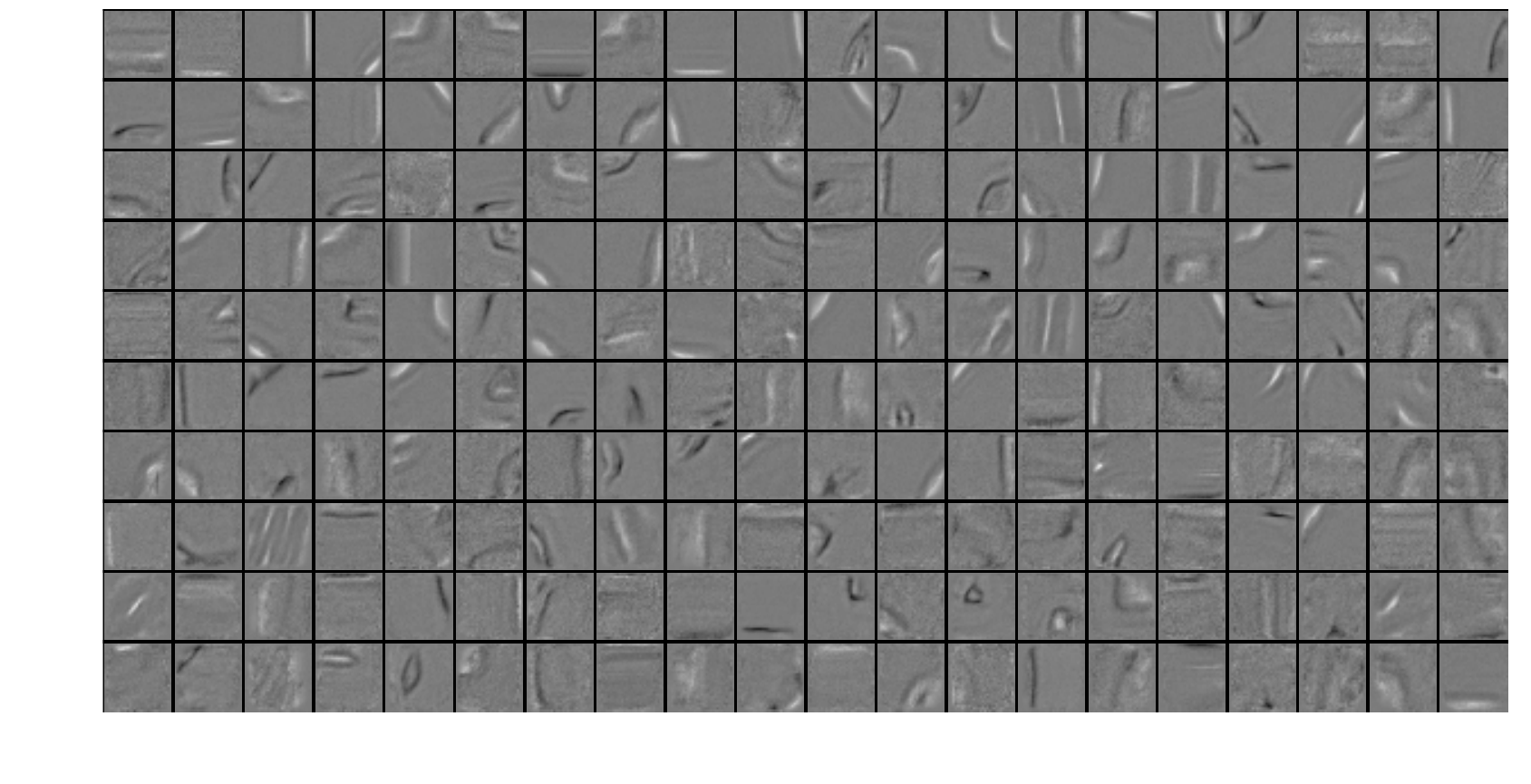}}
  \subfigure[\small Output Gates]{
  \includegraphics[clip=true, trim=0 0 0 100,width=0.48\textwidth]{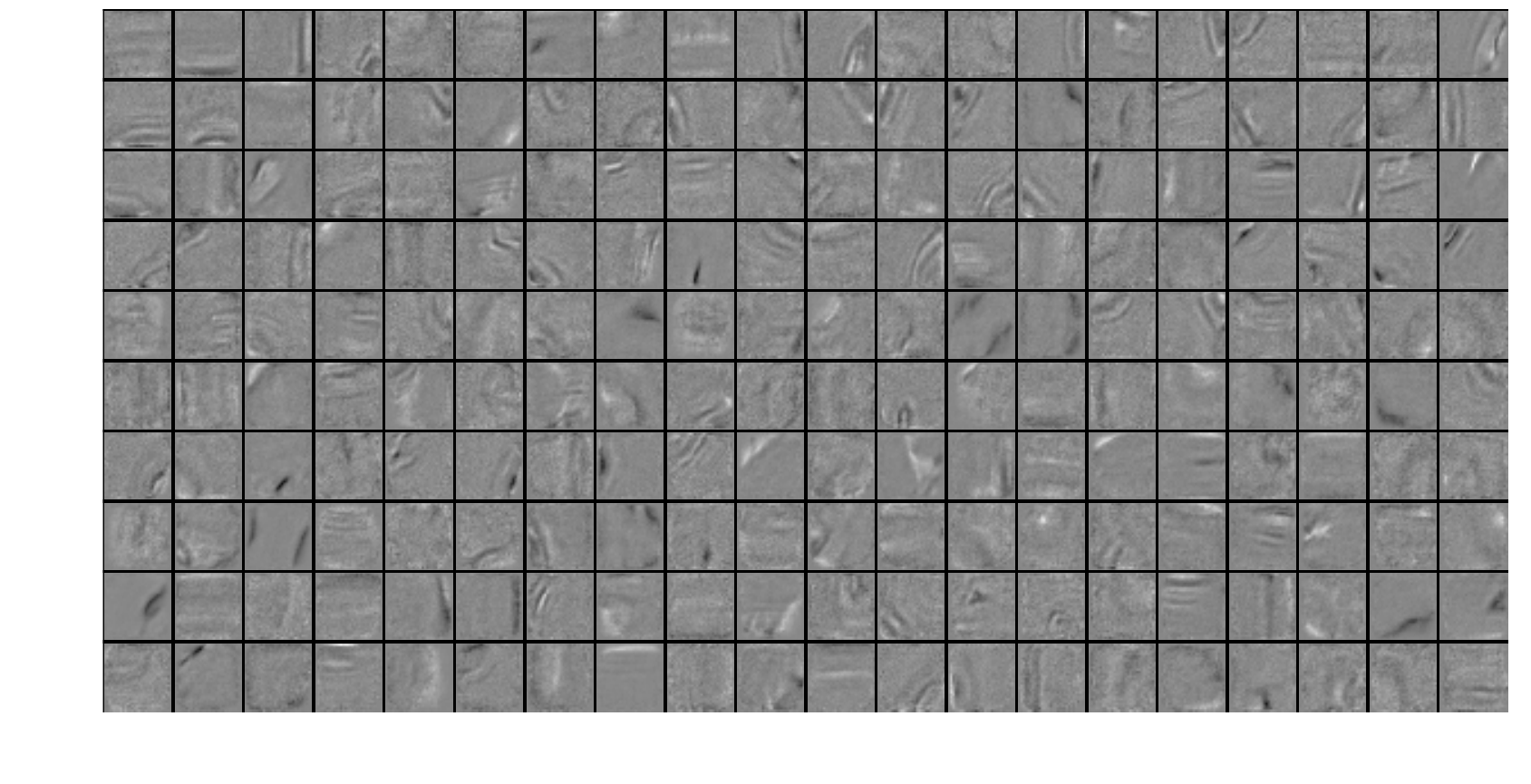}}
  \caption{\small Input features from a Composite Model trained on moving MNIST
  digits. In an LSTM, each input frame is connected to four sets of units - the input, the input
  gate, forget gate and output gate. These figures show the top-200 features ordered by $L_2$ norm of the
input features. The features in corresponding locations belong to the same LSTM unit.}
  \label{fig:input_features}
\end{figure*}

\emph{Visualizing Features} \\
Next, we visualize the features learned by this model.
\Figref{fig:input_features} shows the weights that connect each input frame to
the encoder LSTM.  There are four sets of weights. One set of weights connects
the frame to the input units. There are three other sets, one corresponding to
each of the three gates (input, forget and output). Each weight has a size of 64
$\times$ 64. A lot of features look like thin strips. Others look like higher
frequency strips. It is conceivable that the high frequency features help in
encoding the direction and velocity of motion.

\begin{figure*}
  \centering
  \small
  \subfigure[\small Input Reconstruction]{
  \includegraphics[width=0.48\textwidth]{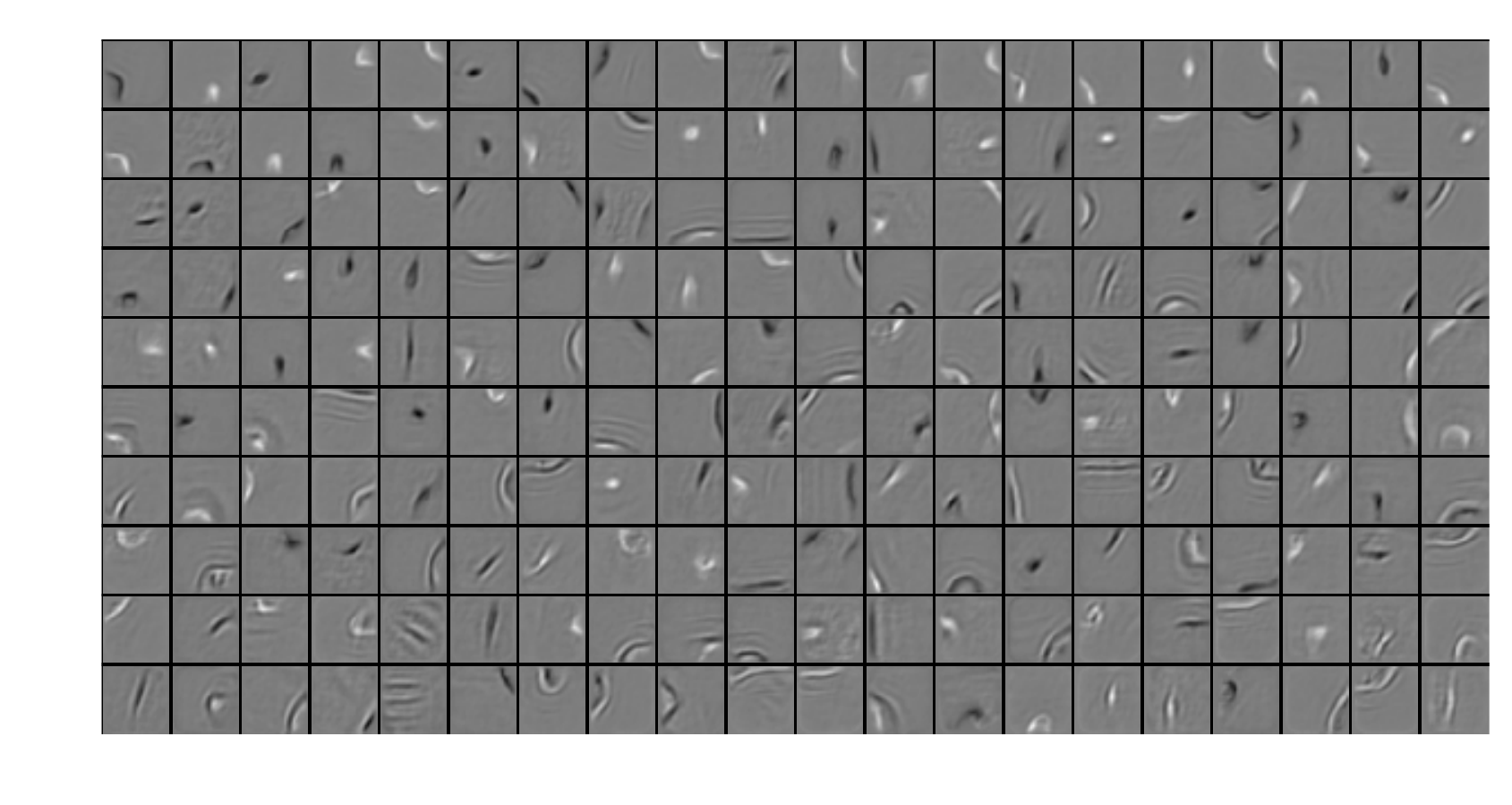}}
  \subfigure[\small Future Prediction]{
  \includegraphics[width=0.48\textwidth]{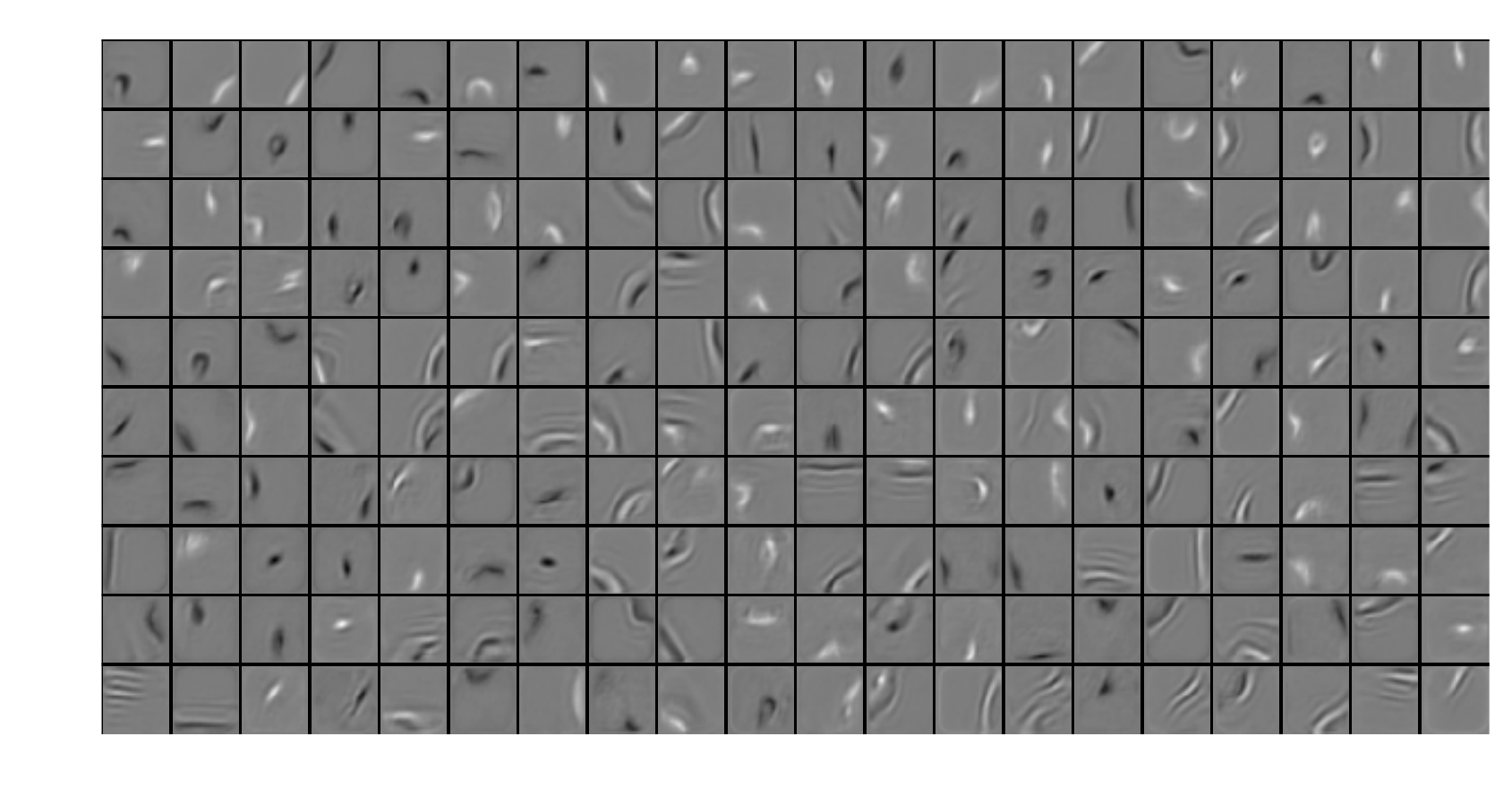}}
  \caption{\small Output features from the two decoder LSTMs of a Composite Model trained on moving MNIST
  digits. These figures show the top-200 features ordered by $L_2$ norm.}
  \label{fig:output_features}
\end{figure*}

\Figref{fig:output_features} shows the output features from the two LSTM
decoders of a Composite Model. These correspond to the weights connecting the
LSTM output units to the output layer. They appear to be somewhat qualitatively
different from the input features shown in \Figref{fig:input_features}. There
are many more output features that are local blobs, whereas those are rare in
the input features. In the output features, the ones that do look like strips
are much shorter than those in the input features. One way to interpret this is
the following. The model needs to know about motion (which direction and how
fast things are moving) from the input. This requires \emph{precise} information
about location (thin strips) and velocity (high frequency strips). But when it
is generating the output, the model wants to hedge its bets so that it 
does not 
suffer a huge loss for predicting things sharply at the wrong place. This could
explain why the output features have somewhat bigger blobs. The relative
shortness of the strips in the output features can be explained by the fact that
in the inputs, it does not hurt to have a longer feature than what is needed to
detect a location because information is coarse-coded through multiple features.
But in the output, the model may not want to put down a feature that is bigger
than any digit because other units will have to conspire to correct for it.


\subsection{Action Recognition on UCF-101/HMDB-51}

The aim of this set of experiments is to see if the features learned by
unsupervised learning can help improve performance on supervised tasks.

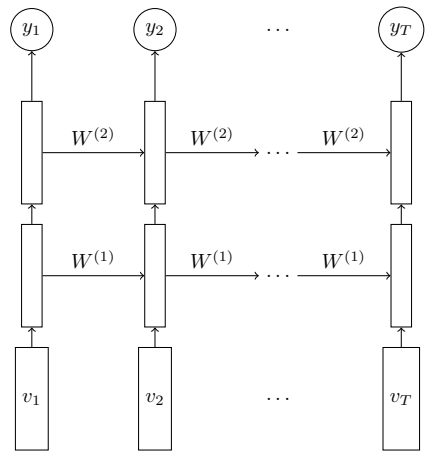
\begin{figure}[ht]
\centering
  \resizebox{0.7\linewidth}{!}{%
    \makeatletter
\ifx\du\undefined
  \newlength{\du}
\fi
\setlength{\du}{\unitlength}
\ifx\spacing\undefined
  \newlength{\spacing}
\fi
\setlength{\spacing}{60\unitlength}
\begin{tikzpicture}
\pgfsetlinewidth{0.5\du}
\pgfsetmiterjoin
\pgfsetbuttcap

\node[rectangle, draw=black, minimum width=10\du, minimum height=50\du] (v1) at (0\spacing, 0) {$v_1$};
\node[rectangle, draw=black, minimum width=10\du, minimum height=50\du] (v2) at (\spacing, 0) {$v_2$};
\node[rectangle, draw=white, minimum width=10\du, minimum height=50\du] (v3) at (2\spacing, 0) {$\ldots$};
\node[rectangle, draw=black, minimum width=10\du, minimum height=50\du] (v4) at (3\spacing, 0) {$v_T$};

\node[rectangle, draw=black, minimum width=10\du, minimum height=50\du] (h1) at (0\spacing, \spacing) {};
\node[rectangle, draw=black, minimum width=10\du, minimum height=50\du] (h2) at (\spacing, \spacing) {};
\node[rectangle, draw=white, minimum width=10\du, minimum height=50\du] (h3) at (2\spacing, \spacing) {$\ldots$};
\node[rectangle, draw=black, minimum width=10\du, minimum height=50\du] (h4) at (3\spacing, \spacing) {};

\node[rectangle, draw=black, minimum width=10\du, minimum height=50\du] (g1) at (0\spacing, 2\spacing) {};
\node[rectangle, draw=black, minimum width=10\du, minimum height=50\du] (g2) at (\spacing, 2\spacing) {};
\node[rectangle, draw=white, minimum width=10\du, minimum height=50\du] (g3) at (2\spacing, 2\spacing) {$\ldots$};
\node[rectangle, draw=black, minimum width=10\du, minimum height=50\du] (g4) at (3\spacing, 2\spacing) {};

\node[circle, draw=black, minimum size=10\du] (y1) at (0\spacing, 3\spacing) {$y_1$};
\node[circle, draw=black, minimum size=10\du] (y2) at (\spacing, 3\spacing){$y_2$};
\node[circle, draw=white, minimum size=10\du] (y3) at (2\spacing, 3\spacing){$\ldots$};
\node[circle, draw=black, minimum size=10\du] (y4) at (3\spacing, 3\spacing){$y_T$};

\draw[->] (v1) -- (h1);
\draw[->] (v2) -- (h2);
\draw[->] (v4) -- (h4);

\draw[->] (h1) -- (g1);
\draw[->] (h2) -- (g2);
\draw[->] (h4) -- (g4);

\draw[->] (g1) -- (y1);
\draw[->] (g2) -- (y2);
\draw[->] (g4) -- (y4);

\draw[->] (h1) -- node[above] {$W^{(1)}$} (h2);
\draw[->] (h2) -- node[above] {$W^{(1)}$} (h3);
\draw[->] (h3) -- node[above] {$W^{(1)}$} (h4);

\draw[->] (g1) -- node[above] {$W^{(2)}$} (g2);
\draw[->] (g2) -- node[above] {$W^{(2)}$} (g3);
\draw[->] (g3) -- node[above] {$W^{(2)}$} (g4);

\end{tikzpicture}
  }
\caption{\small LSTM Classifier.}
\label{fig:lstm_classifier}
\vspace{-0.2in}
\end{figure}

We trained a two layer Composite Model with 2048 hidden units with no conditioning on
either decoders. The model was trained on percepts extracted from 300 hours of
YouTube data. The model was trained to autoencode 16 frames and predict the
next 13 frames. We initialize an LSTM classifier with the
weights learned by the encoder LSTM from this model. The classifier is shown in
\Figref{fig:lstm_classifier}. The output from each LSTM in the second layer goes into a softmax
classifier that makes a prediction about the action being performed at each time
step. Since only one action is being performed in each video in the datasets we
consider, the target is the same at each time step. At test time, the
predictions made at each time step are averaged.  To get a prediction for the
entire video, we average the predictions from all 16 frame blocks in the video
with a stride of 8 frames. Using a smaller stride did not improve results.

The baseline for comparing these models is an identical LSTM classifier but with
randomly initialized weights.  All classifiers used dropout regularization,
where we dropped activations as they were communicated across layers but not
through time within the same LSTM as proposed in \citet{dropoutLSTM}. We
emphasize that this is a very strong baseline and does significantly better than
just using single frames. Using dropout was crucial in order to train good
baseline models especially with very few training examples.

\begin{figure*}[ht]
\centering
\subfigure[UCF-101 RGB]{
\includegraphics[width=0.38\linewidth]{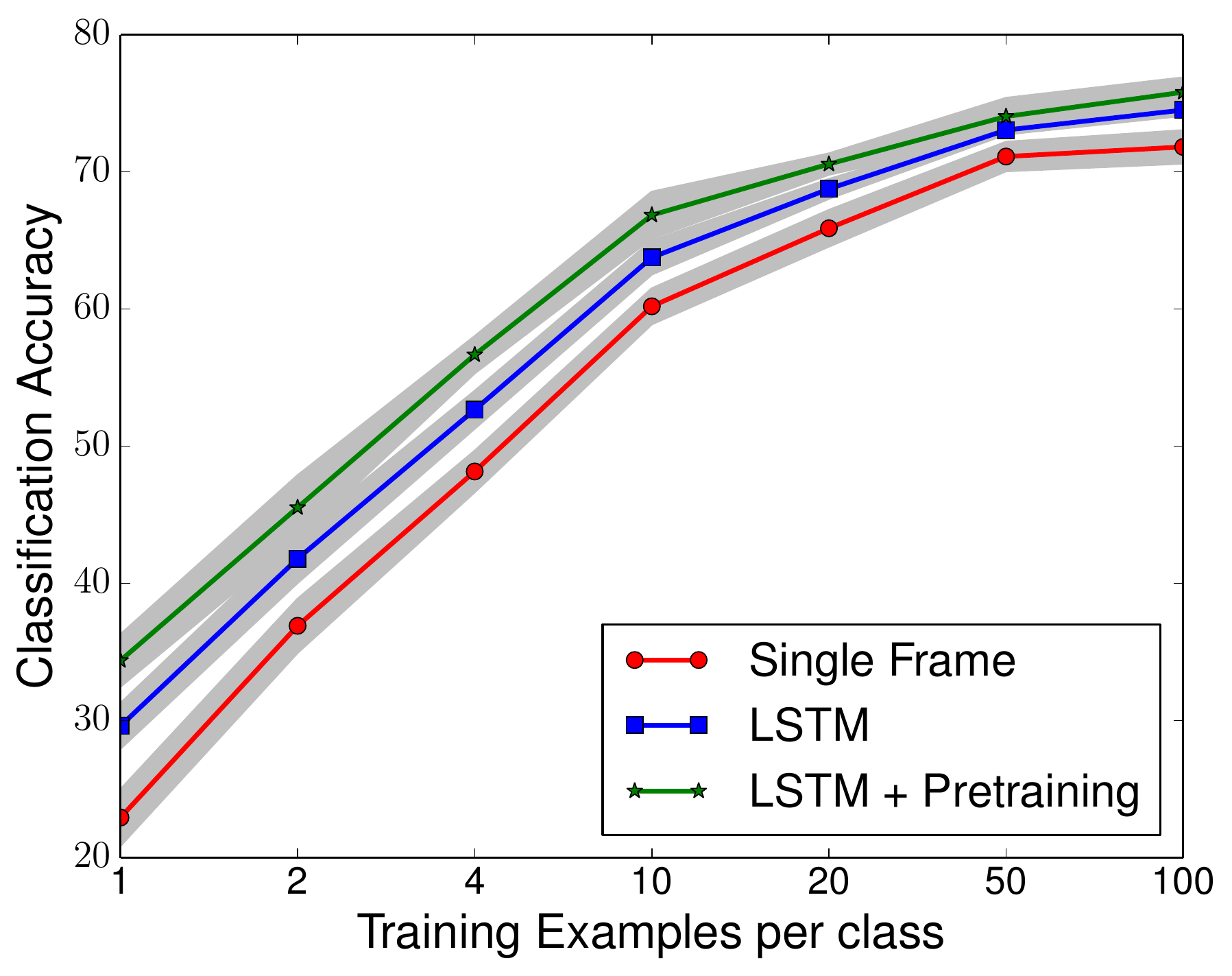}
}
\subfigure[HMDB-51 RGB]{
\includegraphics[width=0.38\linewidth]{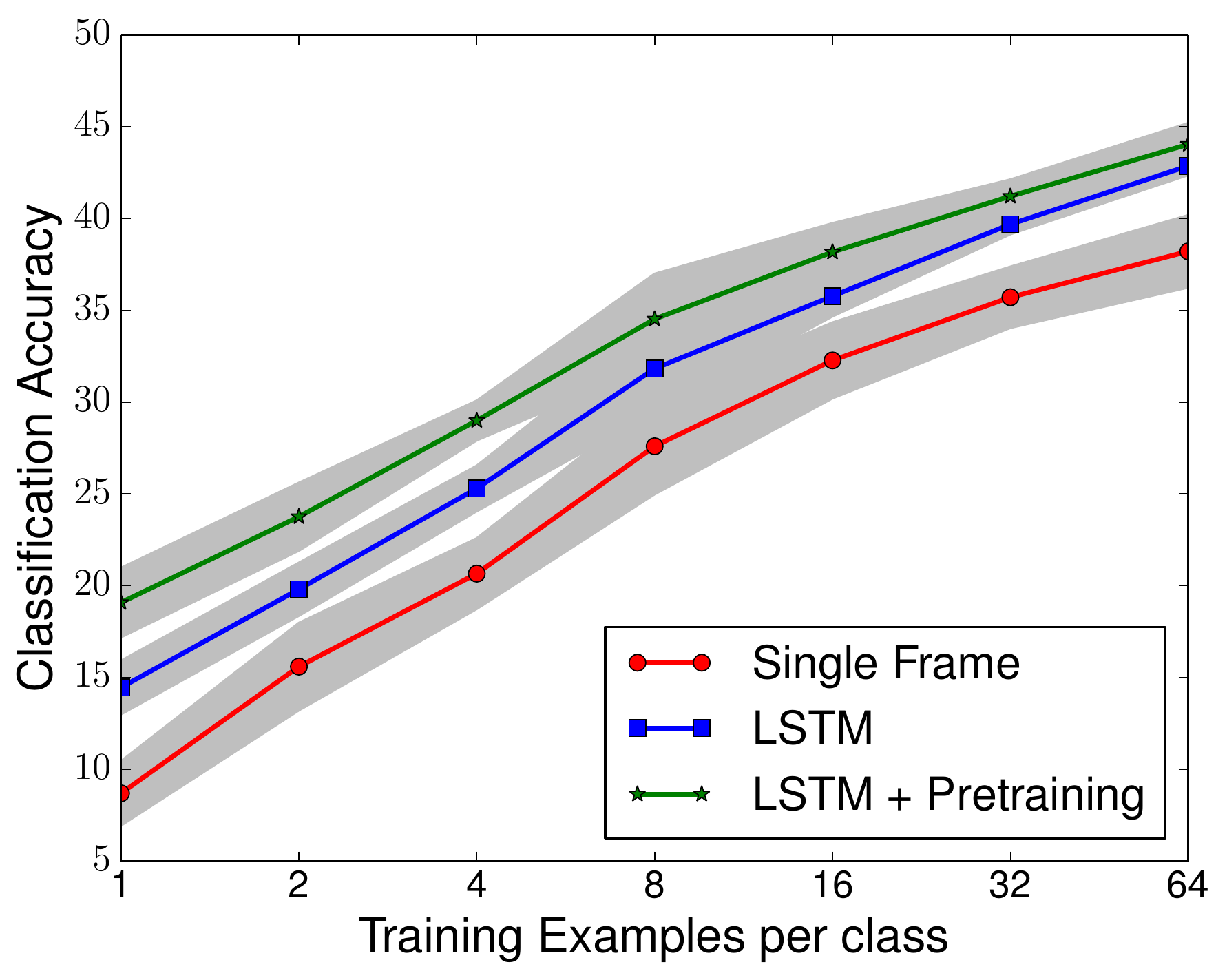}
}
\vspace{-0.1in}
\caption{\small Effect of pretraining on action recognition with change in the size of the labelled training set. The error bars are over 10 different samples of training sets.}
\label{fig:small_dataset}
\vspace{-0.1in}
\end{figure*}

\Figref{fig:small_dataset} compares three models - single frame classifier
(logistic regression), baseline LSTM classifier and the LSTM classifier
initialized with weights from the Composite Model as the number of labelled
videos per class is varied. Note that having one labelled video means having
many labelled 16 frame blocks.  We can see that for the case of very few
training examples, unsupervised learning gives a substantial improvement. For
example, for UCF-101, the performance improves from 29.6\% to 34.3\% when
training on only one labelled video. As the size of the labelled dataset grows,
the improvement becomes smaller. Even for the full UCF-101 dataset we still get a
considerable improvement from 74.5\% to 75.8\%. On HMDB-51, the improvement is
from 42.8\% to 44.0\% for the full dataset (70 videos per class) and 14.4\% to
19.1\% for one video per class.  Although, the improvement in classification by
using unsupervised learning was not as big as we expected, we still managed to
yield an additional improvement over a strong baseline. We discuss some avenues
for improvements later.

We further ran similar experiments on the optical flow percepts extracted from
the UCF-101 dataset. A temporal stream convolutional net, similar to the one
proposed by \citet{Simonyan14c}, was trained on single frame optical flows as
well as on stacks of 10 optical flows. This gave an accuracy of 72.2\% and
77.5\% respectively. Here again, our models took 16 frames as input,
reconstructed them and predicted 13 frames into the future. LSTMs with 128
hidden units improved the accuracy by 2.1\% to 74.3\% for the single frame
case.  Bigger LSTMs did not improve results. By pretraining the LSTM, we were
able to further improve the classification to 74.9\% ($\pm 0.1$). For stacks of
10 frames we improved very slightly to 77.7\%. These results are summarized in
\Tabref{tab:action}.



\begin{table}
\small
\centering
\tabcolsep=0.0in
\begin{tabular}{L{0.35\linewidth}C{0.15\linewidth}C{0.25\linewidth}C{0.25\linewidth}}
\toprule
{\bf Model}         &  {\bf UCF-101 RGB} & {\bf UCF-101 1-~frame flow} & {\bf HMDB-51 RGB}\\
\midrule
Single Frame        &  72.2               & 72.2  & 40.1 \\
LSTM classifier     &  74.5               & 74.3  & 42.8 \\
Composite LSTM Model + Finetuning  &  \textbf{75.8}   & \textbf{74.9}  & \textbf{44.1} \\
\bottomrule
\end{tabular}
\caption{\small Summary of Results on Action Recognition.}
\label{tab:action}
\vspace{-0.2in}
\end{table}

\subsection{Comparison of Different Model Variants}
\begin{table}[t!]
\small
\centering
\tabcolsep=0.0in
\begin{tabular}{L{0.5\linewidth}C{0.25\linewidth}C{0.25\linewidth}} \toprule
{\bf Model}   & {\bf Cross Entropy on MNIST} & {\bf Squared loss on image
patches} \\ \midrule
Future Predictor  & 350.2 & 225.2\\
Composite Model   & 344.9 & 210.7 \\
Conditional Future Predictor & 343.5 & 221.3 \\
Composite Model with Conditional Future Predictor & 341.2   & 208.1 \\
\bottomrule
\end{tabular}
\caption{\small Future prediction results on MNIST and image patches. All models use 2
layers of LSTMs.}
\label{tab:prediction}
\vspace{-0.25in}
\end{table}

\begin{table*}[t]
\small
\centering
\tabcolsep=0.0in
\begin{tabular}{L{0.4\textwidth}C{0.15\textwidth}C{0.15\textwidth}C{0.15\textwidth}C{0.15\textwidth}}\toprule
  {\bf Method}  & {\bf UCF-101 small} & {\bf UCF-101} & {\bf HMDB-51 small} & {\bf HMDB-51}\\\midrule
  Baseline LSTM                                     & 63.7       & 74.5       & 25.3       & 42.8 \\
  Autoencoder                                       & 66.2       & 75.1       & 28.6       & 44.0 \\
  Future Predictor                                  & 64.9       & 74.9       & 27.3       & 43.1 \\
  Conditional Autoencoder                           & 65.8       & 74.8       & 27.9       & 43.1 \\
  Conditional Future Predictor                      & 65.1       & 74.9       & 27.4       & 43.4 \\
  Composite Model                                   & 67.0       & {\bf 75.8} & 29.1       & {\bf 44.1}\\
  Composite Model with Conditional Future Predictor & {\bf 67.1} & {\bf 75.8} & {\bf 29.2} & 44.0 \\ \bottomrule
\end{tabular}
\caption{\small Comparison of different unsupervised pretraining methods. UCF-101 small
is a subset containing 10 videos per
class. HMDB-51 small contains 4 videos per class.}
\label{tab:variants}
\vspace{-0.2in}
\end{table*}

The aim of this set of experiments is to compare the different variants of the
model proposed in this paper. Since it is always possible to get lower
reconstruction error by copying the inputs, we cannot use input reconstruction
error as a measure of how good a model is doing. However, we can use the error
in predicting the future as a reasonable measure of how good the model is
doing. Besides, we can use the performance on supervised tasks as a proxy for
how good the unsupervised model is doing. In this section, we present results from
these two analyses.

Future prediction results are summarized in \Tabref{tab:prediction}. For MNIST
we compute the cross entropy of the predictions with respect to the ground
truth, both of which are 64 $\times$ 64 patches. For natural image patches, we
compute the squared loss. We see that the Composite Model always does a better
job of predicting the future compared to the Future Predictor. This indicates
that having the autoencoder along with the future predictor to force the model
to remember more about the inputs actually helps predict the future better.
Next, we can compare each model with its conditional variant. Here, we find that
the conditional models perform better, as was also noted in \Figref{fig:bouncing_mnist}.

Next, we compare the models using performance on a supervised task.
\Tabref{tab:variants} shows the performance on action
recognition achieved by finetuning different unsupervised learning models.
Besides running the experiments on the full UCF-101 and HMDB-51 datasets, we also ran the
experiments on small subsets of these to better highlight the case where we have
very few training examples. We find that all unsupervised models improve over the
baseline LSTM which is itself well-regularized by using dropout. The Autoencoder
model seems to perform consistently better than the Future Predictor. The
Composite model which combines the two does better than either one alone.
Conditioning on the generated inputs does not seem to give a clear
advantage over not doing so. The Composite Model with a conditional future
predictor works the best, although its performance is almost same as that of the
Composite Model.

\subsection{Comparison with Other Action Recognition Benchmarks}

Finally, we compare our models to the state-of-the-art action recognition
results. The performance is summarized in \Tabref{tab:sota}. The table is
divided into three sets. The first set compares models that use only RGB data
(single or multiple frames). The second set compares models that use explicitly
computed flow features only. Models in the third set use both.

On RGB data, our model performs at par with the best deep models.  It performs
3\% better than the LRCN model that also used LSTMs on top of convnet features\footnote{However, 
the improvement is only partially from unsupervised learning, since we
used a better convnet model.}. Our model performs better than C3D features that
use a 3D convolutional net.  However, when the C3D features are concatenated
with fc6 percepts, they do slightly better than our model.

The improvement for flow features over using a randomly initialized LSTM network
is quite small. We believe this is atleast partly due to the fact that the flow percepts
already capture a lot of the motion information that the LSTM would otherwise
discover.

When we combine predictions from the RGB and flow models, we obtain 84.3
accuracy on UCF-101. We believe further improvements can be made by running the
model over different patch locations and mirroring the patches. Also, our model
can be applied deeper inside the convnet instead of just at the top-level. That
can potentially lead to further improvements.
In this paper, we focus on showing that unsupervised training helps
consistently across both datasets and across different sized training sets.
%

\begin{table}[t]
\small
\centering
\tabcolsep=0.0in
\begin{tabular}{L{0.7\linewidth}C{0.15\linewidth}C{0.15\linewidth}}\toprule
  {\bf Method}                                                  & {\bf UCF-101} & {\bf HMDB-51}\\\midrule
  Spatial Convolutional Net \citep{Simonyan14b}           & 73.0 & 40.5 \\
  C3D \citep{C3D}                                               & 72.3 & - \\ 
  C3D + fc6 \citep{C3D}                                         & {\bf 76.4} & - \\ 
  LRCN \citep{BerkeleyVideo}                             & 71.1 & - \\
  Composite LSTM Model                                    & 75.8 & 44.0 \\
\midrule
  
  Temporal Convolutional Net  \citep{Simonyan14b}         & {\bf 83.7} & 54.6 \\
  LRCN \citep{BerkeleyVideo}                             & 77.0 & - \\
  Composite LSTM Model                                   & 77.7 & - \\
\midrule
  
  LRCN \cite{BerkeleyVideo}                        & 82.9 & - \\
  Two-stream Convolutional Net \cite{Simonyan14b} & 88.0 & 59.4 \\
  Multi-skip feature stacking \cite{MFS}                       & {\bf 89.1} & {\bf 65.1} \\
  Composite LSTM Model                             & 84.3  & - \\
\bottomrule
\end{tabular}
\caption{\small Comparison with state-of-the-art action recognition models.}
\label{tab:sota}
\vspace{-0.2in}
\end{table}

\section{Conclusions} 
\vspace{-0.1in}
We proposed models based on LSTMs that can learn good video representations. We
compared them and analyzed their properties through visualizations. Moreover, we
managed to get an improvement on supervised tasks. The best performing model was
the Composite Model that combined an autoencoder and a future predictor.
Conditioning on generated outputs did not have a significant impact on the
performance for supervised tasks, however it made the future predictions look
slightly better. The model was able to persistently generate motion well beyond
the time scales it was trained for. However, it lost the precise object features
rapidly after the training time scale. The features at the input and output
layers were found to have some interesting properties.

To further get improvements for supervised tasks, we believe that the model can
be extended by applying it convolutionally across patches of the video and
stacking multiple layers of such models. Applying this model in the lower layers
of a convolutional net could help extract motion information that would
otherwise be lost across max-pooling layers. In our future work, we plan to
build models based on these autoencoders from the bottom up instead of applying
them only to percepts.

\vspace{-0.12in}
\section*{Acknowledgments}
\vspace{-0.12in}

We acknowledge the support of Samsung, Raytheon BBN Technologies, and NVIDIA
Corporation for the donation of a GPU used for this research.  The authors
would like to thank Geoffrey Hinton and Ilya Sutskever for helpful discussions
and comments.

\vspace{-0.2in}

{\small 
\bibliography{refs}
\bibliographystyle{icml2015}
}

\end{document}